%% file: main.tex
\documentclass{article}
\newif\ifcameraready
\camerareadytrue

\ifcameraready
\usepackage[final,main]{neurips_2026}
\else
\usepackage[preprint]{neurips_2026}
\fi

\usepackage[utf8]{inputenc}
\usepackage[T1]{fontenc}
\usepackage{courier}  
\usepackage{hyperref}
\hypersetup{colorlinks=true, linkcolor=black, citecolor=black, urlcolor=blue}
\usepackage{url}
\usepackage{graphicx}
\usepackage{booktabs}
\usepackage{xcolor}
\usepackage{xspace}
\usepackage{subcaption}
\usepackage{multirow}
\usepackage{wrapfig}
\usepackage{amsfonts}
\usepackage{amssymb}
\usepackage{amsmath}
\usepackage{nicefrac}
\usepackage{microtype}

\newcommand{\pizhalf}{$\pi_{0.5}$\xspace}

\title{Not All Features Are Created Equal:\\ A Mechanistic Study of Vision-Language-Action Models}

\author{Bryce Grant\thanks{Equal contribution.} \quad Xijia Zhao$^*$ \quad Peng Wang\thanks{Corresponding author: \texttt{pxw206@case.edu}} \\
  Case Western Reserve University \\
  \texttt{\{bag100, xxz1277, pxw206\}@case.edu} \\[2pt]
  \url{https://cwru-aism.github.io/vla-interp-page/}
}

\begin{document}

\maketitle

\begin{abstract}
A fine-tuned VLA policy places the alphabet soup into the basket on command, if given one, just as was demonstrated during fine-tuning. However, when we move the basket five centimeters to the left, it places the soup on the floor off the table. To understand what the policy is thinking, we developed an interpretability pipeline. Guiding the policy with task A's activations by injecting them into the action expert as the policy observes scene B causes \pizhalf{} to reproduce task A's motor trajectory during 99.6\% of episodes ($n{=}1{,}968$). However, this override effect is not universal and it diminishes to 52.1\% on SmolVLA where the interleaved fusion attenuates it, and to 57.0\% on GR00T~N1.5. An injected episode lacks understanding and awareness of the visible scene and instead, it is bound directly to absolute workspace coordinates. This provides the mechanistic evidence explaining the brittleness of VLAs to perturbations that concurrent benchmarks observe. Binding scene to action also holds when deploying to a physical UR10e arm: directed replay caused the arm to replan to the commanded task instead of the prompted one during 26/26 rollouts in one orientation and 18/26 rollouts in another, compared to failing every time in the no-injection and random-direction baselines. Once we physically move the source object used during recording 10--15\,cm away from its expected position, the arm grasps once, then releases at that recorded position in 50/50 of rollouts, picking up the wrong block if another happens to be there or closing its hand at empty air if none does. Systematically applying these interventions to five VLAs (\pizhalf{}, OpenVLA-OFT, X-VLA, SmolVLA, GR00T~N1.5), and ACT as a language-free baseline, across LIBERO, MetaWorld, SimplerEnv, and ALOHA environments and over 420{,}000+ rollouts yields three results: all architectures encode the provided instruction, and four of the five architectures weight it subordinately to vision whenever the goal can be discerned from the scene alone; given standard replacement write-back the specific SAE pooling operation that allows for preserved behavior is specific to architecture (\pizhalf{} uses a per-token pooling while SmolVLA uses one that is indifferent); and finally goal selection and motor generation decomposes into VLM and expert streams wherever the two are made available. We have developed a platform to allow for reproducibility and analysis of our results across all of our trained SAEs, concepts, and rollouts: \textbf{Action Atlas} (\url{https://action-atlas.com}).  Code: \url{https://github.com/CWRU-AISM/action-atlas}.  
\end{abstract}

\section{Introduction}
\label{sec:intro}

Vision-Language-Action (VLA) policies~\citep{brohan2023rt2,kim2024openvla,black2024pi0,pi05,physicalintelligence2025pistar06} fine-tuned on LIBERO~\citep{liu2023libero}, a simulation benchmark for robots, achieve over 90\% success rates (Tab.~\ref{tab:null-injection}). Perturb the object's position by a few centimeters and the policy fails~\citep{zhou2025liberopro,fei2025liberoplus} while the instruction stays the same and the object is visible. Are VLAs following instructions and picking out the correct object? Or are they replaying a motor program bound to where the object sat in their training distribution? If a VLA-driven robot fails, there is no principled way to understand why other than observing its behavior, whereas in traditional robotics the kinematics and control models can be inspected by construction~\citep{craig2018introduction}. 

Sparse Autoencoders (SAEs) allow us to extract sparse monosemantic features from the polysemantic activations of language models~\citep{cunningham2023sparse,bricken2023monosemanticity,templeton2024scaling}. They have been used to find safety-relevant features at scale~\citep{templeton2024scaling} and can be used to steer activations without retraining~\citep{turner2023activation,rimsky2024steering}. There have been multiple concurrent works that have applied these techniques to VLAs (Tab.~\ref{tab:concurrent-work}). \citet{haon2025mech} and \citet{buurmeijer2026observing} steer scalar values like speed and height. \citet{swann2026sparse} and \citet{jin2026event} learn SAEs on a single policy and steer individual features. \citet{miao2026coast} raises success along a learned subspace. All of these works edit one direction on one model with the task held fixed. None of them address the question of whether the policy is acting on the instruction and object or if it is replaying a trajectory toward a location. Steering one feature will not allow us to distinguish between these two, because the object and its training-time location co-vary in every learned feature.

We replace the computation wholesale to see if we can distinguish between these two (Fig.~\ref{fig:system-arch}). We collect activations from a policy while it is doing task A in scene A, then we inject those activations into the policy while it is looking at scene B and told to do task B. If the policy has a concept of the task it should be able to use scene B and move to object B. However, if it is replaying coordinates it will move along its path from task A. We test this across 5 VLAs (\pizhalf{}~\citep{pi05}, OpenVLA-OFT~\citep{kim2025openvlaoft}, X-VLA~\citep{zheng2025xvla}, SmolVLA~\citep{shukor2025smolvla}, GR00T N1.5~\citep{bjorck2025groot}) and 1 language-free control (ACT~\citep{zhao2023learning}) on 4 different benchmarks with over 420,000 rollouts. The models range from 80M to 7B parameters and use flow matching, diffusion, continuous regression and a conditional variational autoencoder (CVAE) to generate actions. When injecting activations from another task, \pizhalf{} moves along the path of the injected task 99.6\% of the time (1,968 episodes) while looking at the destination scene (Fig.~\ref{fig:displacement}). We show the rate per model and label which metric was used to calculate it (ties included) in Tab.~\ref{tab:tie-inclusive}. We denote this as a \emph{coordinate-bound motor program}, an activation-level computation that will replay the path of the injected task, anchored to workspace coordinates, when given a new scene. This gives a mechanistic account of the brittleness reported by \citet{zhou2025liberopro}.

Three more patterns hold across the models we test. (i) There is task-dependent language sensitivity: prompts will affect less behavior if the vision modality knows what the task is, however we find they are easily decodable from the residual stream via linear probing (\S\ref{sec:language-sensitivity}). 

(ii) Specialization in the pathway between goals being selected within the VLM pathways and actions generated by experts when distinct. This is demonstrated on SmolVLA and similar behavior can be seen on \pizhalf{} and GR00T. As well as SmolVLA being resistant to cross-task overriding when testing on LIBERO and MetaWorld due to its interleave of fusion layers (\S\ref{sec:pathway-specialization}). 

(iii) Whether or not SAEs benefit from mean pooling over their features varies by architecture. We find that with our default replace method of writing the SAE reconstruction back into the model \pizhalf{} and OFT need to maintain per token information and SmolVLA does not care. X-VLA and the GR00T Diffusion Transformer (DiT) allow for mean pooling as long as the reconstruction is written back as a residual. When we add an arm to contrast temporally adjacent codes we see degradation on all models. (\S\ref{sec:sae-analysis}).  

Our contributions include:
\begin{itemize}
    \item An analysis of multiple architectures using mechanistic interpretability techniques on four different benchmarks. Each finding is specific to each model tested.
    \item Show that there is a mechanism behind the models creating coordinates that align with a specific task. We use both cross task (\S\ref{sec:cross-task}) and pathway specialization (\S\ref{sec:pathway-specialization}) to find this and show that it is the cause for findings of brittleness on robustness benchmarks. We also replicate this behavior on a physical UR10e robot (\S\ref{sec:physical-body}). 
    \item Findings on using SAEs on VLA models. We test three arms of SAEs (per-token, mean pool and temporal contrastive) and show that different models like different types of SAEs. With the temporal arm we see adjacent code cos similarity rise to ~1 on all models ($\cos_{\text{adj}}{:} 0.86 \to 0.997$). 
    \item Creation of Action Atlas (\url{https://action-atlas.com}).  An open source tool to explore all 520+ SAEs we trained and their 79 concepts we found and over 10{,}000 videos. 
\end{itemize}

\section{Related Work}
\label{sec:related}

RT-2~\citep{brohan2023rt2} introduced VLAs by fine-tuning a VLM to output robot actions as discrete tokens. OpenVLA-OFT~\citep{kim2025openvlaoft} replaced the discrete tokens with continuous actions and reached 97.1\% success on LIBERO. \pizhalf{}~\citep{pi05} adds a separate action expert that generates actions by flow matching. The VLAs in this study connect the VLM to the action generator in different ways (Tab.~\ref{tab:architectures}). For example, the \pizhalf{} expert attends to the VLM tokens at every layer, and the SmolVLA~\citep{shukor2025smolvla} expert alternates cross-attention to the VLM with self-attention over the action tokens.

The performance of current VLAs drops sharply with small perturbations~\citep{fei2025liberoplus}. Moving objects slightly causes OpenVLA~\citep{kim2024openvla} and $\pi_0$~\citep{black2024pi0} to fail at 0\% success on all LIBERO tasks, and \pizhalf{} reaches at most 0.38~\citep{zhou2025liberopro}. Also, an instruction for an untrained task reduces success to $\sim$0\% on every suite.

Mechanistic interpretability offers tools to look for the cause of these failures inside the policy. Activation patching copies activations from one forward pass into another to find the components that carry a behavior in language models~\citep{meng2022locating}. SAEs~\citep{bricken2023monosemanticity,cunningham2023sparse,templeton2024scaling} and steering vectors~\citep{turner2023activation,rimsky2024steering} also come from language models, and the properties of VLAs change their use. Action tokens take phase structure that text tokens lack, so per-token and mean-pooled SAEs perform differently (\S\ref{sec:sae-analysis}). And discrete action tokens turn small reconstruction errors into wrong action bins, therefore SAE insertion reduces base OpenVLA to 0\% success at every layer except the last (App.~\ref{sec:discrete-vla}).

Several studies edit or probe the activations of VLAs. Interpolating the averaged text-token hidden states of two tasks in $\pi_0$ chains their behaviors into a new task, and $\pi_0$ maps object names to their demonstrated locations~\citep{li2025textlatent}. Clamping feed-forward neurons in OpenVLA and $\pi_0$-FAST changes speed and lift height~\citep{haon2025mech}.  On DROID hardware, SAE features of \pizhalf{} steer object choice under an empty prompt~\citep{swann2026sparse}, and a minimal-norm linear controller on top of linear observers changes the grasp and the chosen object of \pizhalf{}~\citep{buurmeijer2026observing}. Gating one layer toward a subspace of successful rollouts raises success on the commanded task in \pizhalf{}, $\pi_0$-FAST and GR00T~N1.5~\citep{miao2026coast}. Suppressing SAE features aligned with behavioral events reduces the success of OpenVLA and \pizhalf{}~\citep{jin2026event}. An SAE trained on text steers the gripper of a VLA~\citep{khan2025sparse}, and probes find world-model structure in OpenVLA~\citep{molinari2025worldmodel}. Tab.~\ref{tab:concurrent-work} summarizes the architectures, interventions and hardware of these studies. See App.~\ref{sec:extended-related} for more related work.

\begin{table}[b]\centering\footnotesize
\caption{VLA-interpretability work summary. Arch.: architectures intervened on. FFN: feed-forward network. X-task: cross-task activation injection. Pool/Path: SAE pooling comparison and VLM-expert pathway split. HW: real robot.}
\label{tab:concurrent-work}
\label{tab:related-delta}
\begin{tabular}{@{}lccccc@{}}
\toprule
\textbf{Work} & \textbf{Arch.} & \textbf{Intervention} & \textbf{X-task} & \textbf{Pool/Path} & \textbf{HW} \\
\midrule
\citet{haon2025mech}          & 2 & FFN, kinematic    & no & no/no & UR5 \\
\citet{swann2026sparse}       & 1 & SAE feature      & no & no/no & DROID \\
\citet{buurmeijer2026observing} & 1 & linear, kinematic + object & no & no/no & DROID \\
\citet{miao2026coast}         & 3 & subspace, success & no & no/no & DROID \\
\citet{jin2026event}          & 2 & SAE, suppression  & no & no/no & ALOHA \\
\citet{khan2025sparse}        & 1 & SAE direction     & no & no/no & sim \\
\textbf{This work} & \textbf{6} & \textbf{injection + SAE} & \textbf{yes} & \textbf{yes/yes} & \textbf{sim + UR10e} \\
\bottomrule
\end{tabular}
\end{table}

\section{Method}
\label{sec:method}
\begin{figure}[t]
    \centering
    \includegraphics[width=\columnwidth]{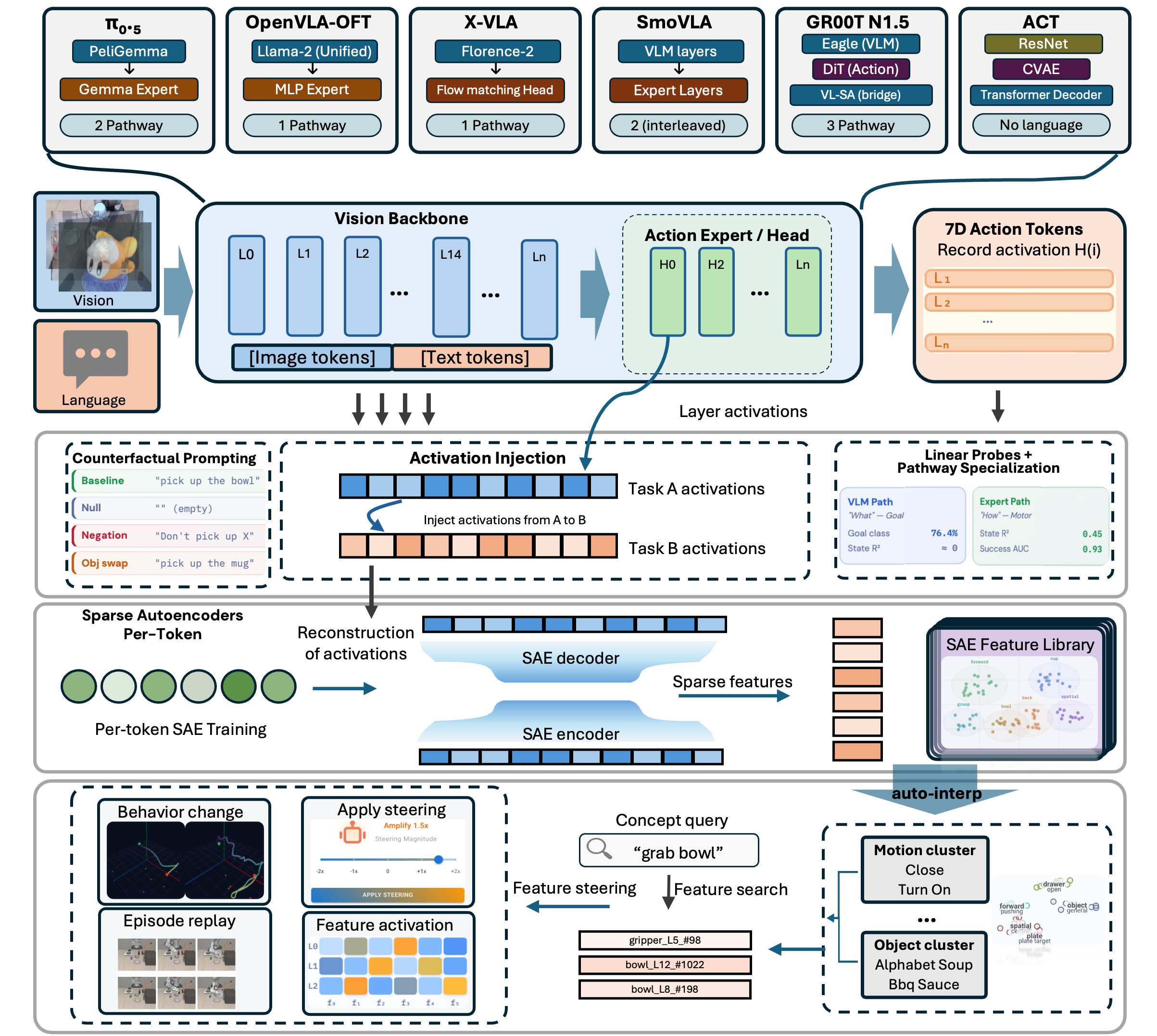}
    \caption{\textbf{Methodology overview.} Top: record a VLA's activations during a successful rollout and replay them under cross-task injection and counterfactual prompts. Middle: per-token SAEs decompose the same activations into features. Bottom: features are clustered, searched, and validated by ablation and steering in Action Atlas.}
    \label{fig:system-arch}
\end{figure}

Our pipeline is closed-loop in that we use rollouts as our ground truth for any claims about mechanism (Fig.~\ref{fig:system-arch}). We can intervene on the policy using activation injection and counterfactual prompts. Both allow us to extract causality from the resulting change in behavior. Activation injection allows us to replay computations within the forward pass and counterfactual prompts allow us to modify the language input while keeping the scene constant. We can then use SAEs to extract features from these saved activations and localize causality to certain layers and subspaces. Linear probes can be used to understand what information the residual stream contains at each layer~\citep{haoxuanSAE}. We validate with the same intervention techniques. Any feature we find with our SAEs counts as a mechanism only if it causes the same effect through injection and rollout.

\subsection{VLA Architectures Under Study}

We use five VLAs and one model without language as a control. These include: \pizhalf{}~\citep{pi05} (3B params, flow matching), OpenVLA-OFT~\citep{kim2025openvlaoft} (7B params, continuous L1 loss), X-VLA~\citep{zheng2025xvla} (1B params, soft-prompted flow), SmolVLA~\citep{shukor2025smolvla} (450M params, VLM-expert interleaved), GR00T N1.5~\citep{bjorck2025groot} (3B params, DiT-Eagle-VL-SA) and ACT~\citep{zhao2023learning} (80M params, CVAE, no language). This covers models from 80M to 7B params and four types of action generation: flow matching~\citep{flowmatching}, continuous regression, diffusion and CVAE decoding (see Tab.~\ref{tab:architectures}).

\subsection{Activation Injection}

\emph{Activation injection} extends activation patching~\citep{meng2022locating} to full rollout episodes: we replace activations from one episode with those from another during inference. Given a successful source episode $A$ and target episode $B$, we record the layer activations $\{\mathbf{H}^{A,(\ell)}\}$ during the rollout of $A$, then at every timestep $t$ of rollout $B$, we override the residual stream at the chosen layers,
\begin{equation}
\mathbf{H}^{B,(\ell)}_t \;\leftarrow\; \mathbf{H}^{A,(\ell)}_t, \qquad \ell \in \mathcal{L}_{\text{inject}},
\label{eq:injection}
\end{equation}
and let the rest of the forward pass proceed. We vary the injection set $\mathcal{L}_{\text{inject}}$ from a single layer to all layers. If $A$ and $B$ produce residual streams with varying sequence lengths, we overwrite the 50 positions within the action chunk and pass through the image and prompt tokens without modification. When the target episode runs longer than the recorded source ($T_B > T_A$), the injection ends with the source length $t{=}T_A$ and the target policy continues unmodified (this impacts $<$5\% of cross-task episodes; restricting to $T_B \leq T_A$ shifts override $\leq$1pp on \pizhalf{} \texttt{libero\_goal}).

We have three injection settings. \textbf{Null injection} will use the correct prompt on the source episode and no prompt on the target. \textbf{Same-scene injection} will use the same scene for both episodes but instruct them to interact with different objects. \textbf{Cross-task injection} will use two different scenes. For models with multiple pathways (\pizhalf{}, SmolVLA, GR00T) we also inject into each pathway individually to isolate their contributions.

\subsection{Counterfactual Prompting}

We change the text prompt to see how much the language affects behavior. Each episode will have one of six prompt conditions: the correct prompt, null (empty string), negation (e.g., ``Don't pick up X''), motor instructions (e.g., ``Move slowly''), object swap (replace the target object's name) and temporal switch (change the prompt partway through the episode). We also use counterfactual prompts on SmolVLA for the different difficulty levels on MetaWorld (easy, medium, hard, very hard).

\subsection{Sparse Autoencoders for VLAs}
\label{sec:sae}

We train SAEs on activations related to actions, using TopK sparsity~\citep{gao2024scaling} with $k{=}64$ and an expansion factor of $m{=}4d$ or $m{=}8d$. It is important to keep token-level information for some models (\S\ref{sec:sae-analysis}). When we take the mean over the 50 action tokens for \pizhalf{} we lose the ability to distinguish between approaching the object, interacting with it and finishing the task. Rollout success goes from 89.9\% to 3.3\% (average over all four LIBERO suites and when sweeping over layers for just \texttt{libero\_object} we see a similar trend, 93.8\% to 8.9\%), despite a normalized MSE $\leq 0.05$ ($R^2 > 0.95$). We report normalized MSE as in \citet{gao2024scaling}. We process each token individually as the conservative default. More on pooling by model is in \S\ref{sec:sae-analysis} and App.~\ref{sec:temporal-sae-arm}. To find concept-specific features we use frequency-weighted contrast ($\text{score}_f = d_f \times \text{freq}_f$). For each feature we take the standardized difference in activation between tasks containing the concept and those that do not. This is the Cohen's $d$ ($d_f$)~\citep{cohen1988statistical} of the feature. We then multiply this by $\text{freq}_f$, the fraction of samples in which the feature is in the top-$k$ active features. We train over 520 SAEs across the five VLAs (see Tab.~\ref{tab:episode-counts} for the number of SAEs per model), finding 79 unique concepts related to manipulation (the union of concept names across models; the per-model lists overlap heavily). These concepts cover different aspects of motion, objects and space.

\subsection{Linear Probes for Action Prediction}
\label{sec:linear-probes}

Linear probes~\citep{alain2017understanding} test whether action information is linearly decodable from intermediate representations. We train ridge regression probes per action dimension and verify causality by projecting out the probe direction.

\subsection{Metrics}

We evaluate three primary metrics. \textbf{Action Cosine Similarity} measures behavioral alignment between episodes, $\cos(\mathbf{a}^X, \mathbf{a}^Y) = \langle \mathbf{a}^X, \mathbf{a}^Y\rangle / (\|\mathbf{a}^X\|\|\mathbf{a}^Y\|)$, computed over the full per-token action vectors. \textbf{Task Success} is a binary indicator determined by the environment's built-in success criteria. We report the change in success rate (SR) as $\Delta\text{SR} = \text{SR}_{\text{intervention}} - \text{SR}_{\text{baseline}}$. \textbf{Override Rate} quantifies how often the injected behavior dominates the prompted behavior in cross-task injection. Let $\Delta_t = \cos(\mathbf{a}^{B|A\text{-inj}}_t, \mathbf{a}^{A}_t) - \cos(\mathbf{a}^{B|A\text{-inj}}_t, \mathbf{a}^{B}_t)$. We define the override rate conditional on exclusion of a cosine-tie band $|\Delta_t| \leq \tau$,
\begin{equation}
\text{OR} \;=\; \Pr\!\left[\,\Delta_t > \tau \mid |\Delta_t| > \tau\,\right],
\label{eq:override}
\end{equation}
with $\tau = 0.05$. The share of steps inside the tie band varies by architecture, from 0.3\% on \pizhalf{} to 60.0\% on X-VLA, so we also report \emph{tie-inclusive} override, $\Pr[\Delta_t > 0]$ over all injection steps, for every model (Tab.~\ref{tab:tie-inclusive}, Fig.~\ref{fig:displacement}).
\section{Experiments}
\label{sec:experiments}

Our results show that the scene determines which motor program is selected (\S\ref{sec:visual-dominance}, \S\ref{sec:cross-task}). Language only plays a role when there is ambiguity in the scene (\S\ref{sec:language-sensitivity}). Different pathways are responsible for different parts of this process. The VLM selects the coordinate-bound motor program and the expert binds it to coordinates (\S\ref{sec:pathway-specialization}). We can use SAEs to find features that represent this information, with a pooling choice that depends on the architecture (\S\ref{sec:sae-analysis}). We test this binding on a real robot arm (\S\ref{sec:physical-body}).

\subsection{Experimental Setup}

We collect 420{,}000+ rollout episodes across 12 experiment types and up to 50 tasks per environment (scale in Table~\ref{tab:episode-counts}), on 8$\times$A100-80GB, RTX 5090, and 2$\times$RTX 4090, over four benchmarks: \textbf{LIBERO}~\citep{liu2023libero} (4 suites, 40 tasks), \textbf{MetaWorld}~\citep{yu2020metaworld} (50 tasks, 4 difficulty levels), \textbf{SimplerEnv}~\citep{simpler_env} (10 tasks, 2 embodiments), and \textbf{ALOHA}~\citep{zhao2023learning} (2 bimanual tasks).

\subsection{Visual Context Dominates Action Generation}
\label{sec:visual-dominance}

Visual context dominates action generation across all six architectures (Tab.~\ref{tab:null-injection}). By injecting the base activation from the visual pathway when using the null prompt we get cosine similarity back to $\geq$ 0.99 and get task success rates of 14--80\%. If we set one layer to zero we see both X-VLA and GR00T-DiT go to 0\%. SmolVLA can survive single expert-layer zeroing (19.2--76.5\% across suites). We found greater success rates when injecting into dual-pathway models (\pizhalf{} had success rates of 70--80\% per suite at $n{=}50$/suite, against  null injection controls of 22\%, 6.0\% and 62\%, Tab.~\ref{tab:null-injection}). And smaller success rates for single-pathway models such as OFT (14--15\%, $n{=}120$ per layer). We believe this is due to language being more involved with the visual pathway in these models as they have to rely on language more to understand the prompt (\S\ref{sec:language-sensitivity}). We find that the model recovers some percentage of the success rate lost by the prompt. It restores 88\% of the lost success on \texttt{libero\_goal} and 69\% on \texttt{libero\_object}, while the \texttt{libero\_spatial} contrast is not significant ($p{=}0.13$), the suite where an empty prompt costs 26\%. Per-model layer-zeroing profiles in App.~\ref{sec:additional-results}.

\begin{table}[ht]
\centering
\caption{Visual-pathway influence per suite. Baselines come from a unified 25-ep$\times$10-task batch ($n{=}250$/suite), and the OFT null prompt is $n{=}50$/suite. $^{\ast}$One matched \pizhalf{} batch ($n{=}50$/suite) with its own baselines of 90.0\%, 92.0\% and 88.0\% (App.~\ref{sec:additional-results}). $^{\ast\ast}$From the $n{=}500$ sweep of Tab.~\ref{tab:counterfactual}. Same-scene: baseline activations injected into a same-task episode. Rec.: OFT null prompt plus injected baseline activations. Zero-layer rows average the layer sweep. $^{\dagger}$The public GR00T spatial checkpoint underperforms the reference (App.~\ref{sec:per-arch-failures}).}
\label{tab:null-injection}
\footnotesize
\resizebox{\columnwidth}{!}{%
\begin{tabular}{@{}llcccc@{}}
\toprule
\textbf{Model} & \textbf{Condition} & \textbf{Goal} & \textbf{Object} & \textbf{Spatial} & \textbf{Long\,/\,10} \\
\midrule
\multirow{4}{*}{\pizhalf{}}
 & Baseline                           & \phantom{0}93.2\%        & \phantom{0}87.6\%        & \phantom{0}92.8\%        & \phantom{0}86.0\%        \\
 & Null prompt, no injection$^{\ast}$ & \phantom{00}6.0\%        & \phantom{0}22.0\%        & \phantom{0}62.0\%        & \phantom{0}56.6\%$^{\ast\ast}$ \\
 & Null + PaliGemma-ALL inject$^{\ast}$ & \phantom{0}80.0\%      & \phantom{0}70.0\%        & \phantom{0}78.0\% & ---                \\
 & Same-scene inject ALL              & \phantom{0}91.0\%        & \phantom{0}87.5\%        & \phantom{0}88.2\%        & \phantom{0}75.2\%        \\
\midrule
\multirow{3}{*}{OFT}
 & Baseline                           & \phantom{0}98.8\%       & \phantom{0}99.2\%       & \phantom{0}92.4\%       & \phantom{0}91.2\%       \\
 & Null prompt, no injection          & \phantom{0}12.0\%       & \phantom{0}96.0\%       & \phantom{0}58.0\%       & ---                     \\
 & Null + inject zero-baseline (rec.) & \phantom{0}14.1\%       & \phantom{0}14.6\%       & \phantom{0}14.4\%       & \phantom{0}13.5\%       \\
\midrule
\multirow{3}{*}{X-VLA}
 & Baseline                           & \phantom{0}98.8\%      & \phantom{0}98.8\%      & \phantom{0}95.6\%      & \phantom{0}87.6\%      \\
 & Null prompt, no injection          & \phantom{0}10.0\%       & \phantom{0}60.0\%       & \phantom{0}48.0\%       & \phantom{0}28.0\%       \\
 & Zero any single layer              & \phantom{00}0.0\%       & \phantom{00}0.0\%       & \phantom{00}0.0\%       & \phantom{00}0.0\%       \\
\midrule
\multirow{2}{*}{SmolVLA}
 & Baseline                           & \phantom{0}74.8\%        & \phantom{0}87.6\%        & \phantom{0}65.6\%        & \phantom{0}41.2\%        \\
 & Zero expert layer (mean)           & \phantom{0}71.0\%             & \phantom{0}76.5\%             & \phantom{0}61.8\%             & \phantom{0}19.2\%   \\
\midrule
\multirow{3}{*}{GR00T N1.5}
 & Baseline                           & \phantom{0}93.6\%      & \phantom{0}95.6\%      & \phantom{0}69.2\%$^{\dagger}$ & \phantom{0}77.2\%      \\
 & Null source (inject ALL DiT)       & \phantom{00}3.3\%       & \phantom{00}3.3\%       & ---$^{\dagger}$ & \phantom{00}3.3\%       \\
 & Zero any DiT layer (L0--L15)       & \phantom{00}0.0\%       & \phantom{00}0.0\%       & ---$^{\dagger}$ & \phantom{00}0.0\%       \\
\midrule
\multirow{2}{*}{ACT (ALOHA)}
 & Baseline ($n{=}100$/task)          & \multicolumn{4}{c}{100.0\% TransferCube, 78.0\% Insertion}                                                                          \\
 & Mask workspace grid (2,2), $n{=}20$ & \multicolumn{4}{c}{\phantom{0}10.0\% TransferCube, 20.0\% Insertion}                                                               \\
\bottomrule
\end{tabular}}
\end{table}

When we inject activations from prompt $B$ into rollouts of prompt $A$ within the same scene we can steer towards the injected target on every multi-pathway model tested. \pizhalf{} also tolerates zeroing out later layers, similar to SmolVLA. Once vision has picked an action, what does the activation represent? See \S\ref{sec:cross-task}.

\subsection{Motor Programs Bind to Absolute Spatial Coordinates}
\label{sec:cross-task}

Is the injected activation representing a concept of the task (``pick up the cup'') or a coordinate-bound motor program? We test this by injecting activations across tasks. Here we overwrite the expert-pathway residual stream at layers 16 and 17 across all 50 action-token positions at every timestep. For all five language-capable VLAs, destination-task success collapses, and trajectory analysis shows behavior steered toward source-task positions (Tab.~\ref{tab:cross-task} with ACT serving as the language-free control).

\begin{table}[ht]
\centering
\caption{Cross-task injection across six models. Dst.\ SR: destination-task success under injection. Override rate: tie-band-conditional at $\tau{=}0.05$ (Eq.~\ref{eq:override}), with tie-inclusive rates in Tab.~\ref{tab:tie-inclusive}. ACT's Dst.\ SR is its uninjected Insertion rate on the same 10 seeds.}
\label{tab:cross-task}
\footnotesize
\begin{tabular}{@{}lrrrl@{}}
\toprule
\textbf{Model} & \textbf{Pairs ($n$)} & \textbf{Dst.\ SR} & \textbf{Override rate} & \textbf{Notes} \\
\midrule
\pizhalf{}  & 1{,}968         & \phantom{0}2.6\%             & 99.6\% & goal, spatial, object \\
OFT         & 1{,}079         & \phantom{0}5.0\%             & 96.4\% & 90.8--100.0\% per suite \\
X-VLA       & 3{,}150         & \phantom{0}0.0\% & 99.8\% & LIBERO; WidowX 54.2\% \\
SmolVLA     & \phantom{0,}720 & \phantom{0}0.0\%             & 52.1\% & expert 78.6\%, VLM 25.6\% \\
GR00T       & \phantom{0,}270 & \phantom{0}0.0\%             & 57.0\% & goal 85.6\%, object 52.2\%, long 33.3\% \\
ACT (ctrl.) & \phantom{0,0}10 & 70.0\%                & \phantom{0}0.0\% & no-op ($\cos{=}1.0$) \\
\bottomrule
\end{tabular}
\end{table}

\begin{figure}[t]
    \centering
    \includegraphics[width=\columnwidth]{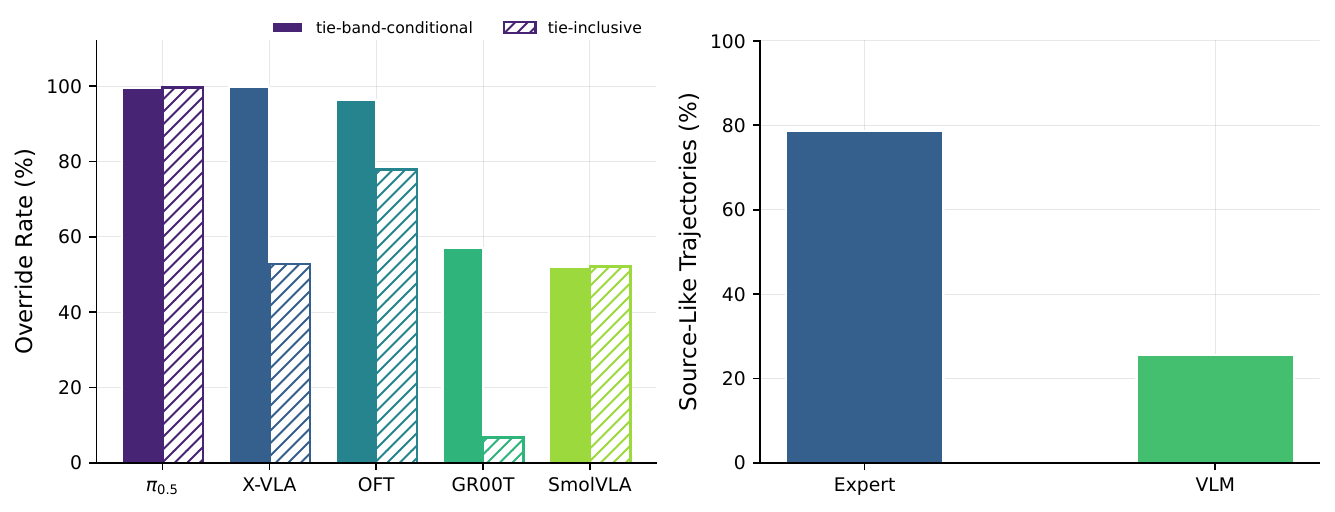}
    \caption{\textbf{Cross-task override rates.} Left: tie-band-conditional (solid) and tie-inclusive (hatched) override per model, with metrics and CIs in Tab.~\ref{tab:tie-inclusive}. Right: SmolVLA override by injected pathway on LIBERO ($n{=}720$).}
    \label{fig:displacement}
\end{figure}

Analyzing the displacement of the injected trajectories (Figure~\ref{fig:displacement}), we find that injection decreases success and still steers behavior. \pizhalf{}, X-VLA, and OFT all show strong source dominance ($\geq$96\% conditional on the tie band, and 52.8--99.6\% when ties are included). GR00T shows moderate source dominance, with 57.0\% under action cosine and 6.7\% under end-effector velocity. SmolVLA's expert pathway achieves a 3.07$\times$ greater override rate than its VLM pathway. Objects do move, but they are the objects of the destination task. Using DTW on the end-effector trajectories, we find source dominance of 86--91\% on \texttt{libero\_goal} and \texttt{libero\_spatial} (App.~\ref{sec:dtw-cross-task}).

A random direction of the same magnitude at the same layers gives chance-level override (App.~\ref{sec:random-injection}), so the effect comes from the injected direction. When we align the source and destination tasks by motion phase, the override rate remains high at 94.4\% (App.~\ref{sec:dtw-cross-task}). Reducing the fine-tuning data by an order of magnitude gives the same result (App.~\ref{sec:datascale}), as does halving the injection strength (App.~\ref{sec:dose-response}). The robot therefore moves to where the source objects would have been, and its motor program is tied to absolute workspace coordinates, which agrees with prior findings~\citep{fei2025liberoplus,zhou2025liberopro}. We also observe this in our physical experiments (\S\ref{sec:physical-body}).

\subsection{Language Becomes Necessary Only When Vision is Ambiguous}
\label{sec:language-sensitivity}

If the visual context determines the action, then language should only play a role when the scene does not provide enough information to identify the task. We test this with counterfactual prompts across five language-capable VLAs (per-model $n$ in App.~\ref{sec:additional-results}) and find that it holds in graded form. Task structure sets the ordering of prompt tolerance. This ordering is consistent across models, but its magnitude changes between models. Across the four LIBERO suites and four levels of MetaWorld difficulty, every model tested on both benchmarks has a higher success rate under a null or incorrect prompt on tasks with clear visual cues (\texttt{libero\_object}, MetaWorld easy) than on visually ambiguous tasks (\texttt{libero\_goal}, MetaWorld hard), by at least a factor of two. \pizhalf{} has a success rate of 31.6\% with a null prompt on \texttt{libero\_object} and 5.8\% on \texttt{libero\_goal}, compared to baselines of 91.0\% and 92.6\% respectively ($n{=}500$/cell). \pizhalf{} does not follow the meaning of instructions. When told \texttt{do not \{task\}}, it completes the task 96.7\% of the time ($n{=}30$, within-run baseline 86.7\%).

The magnitude changes between models and spans the entire range on one suite. On \texttt{libero\_object}, OFT maintains a success rate of 96.0\% when given a null prompt ($n{=}50$/cell), \pizhalf{} falls to 31.6\%, and SmolVLA is at 0.0\% (\pizhalf{} and SmolVLA $n{=}500$/cell, baselines of 100.0\%, 91.0\% and 89.2\% respectively). The SmolVLA checkpoint for LIBERO either succeeds or fails completely. It succeeds in 0/500 episodes with a null prompt and in 0/12 with an empty, a one-token, or a nuisance prompt, compared to 11/12 with the true instruction. The SmolVLA checkpoint for MetaWorld, however, maintains a success rate of 82.0\% with a null prompt on the easy difficulty (Tab.~\ref{tab:counterfactual}).

\begin{table}[ht]
\centering
\caption{Counterfactual prompting across models ($n$ per model in App.~\ref{sec:additional-results}). The $n{=}500$ \pizhalf{} and SmolVLA cells are 50 episodes per task from one fixed init state (LeRobot 0.4.4~\citep{cadene2024lerobot} pins it), paired on layout across conditions. $^{\ddagger}$Matched within-experiment control, within 6\,pp of the Tab.~\ref{tab:null-injection} baselines. $^{\S}$Not run on MetaWorld.}
\label{tab:counterfactual}
\footnotesize
\begin{tabular}{@{}lccc@{}}
\toprule
\textbf{Model / Suite} & \textbf{Correct Prompt}$^{\ddagger}$ & \textbf{Null} & \textbf{Wrong Obj.}$^{\S}$ \\
\midrule
\pizhalf{} / object  & \phantom{0}91.0\%   & \phantom{0}31.6\%   & ---                 \\
\pizhalf{} / goal    & \phantom{0}92.6\%   & \phantom{00}5.8\%   & ---                 \\
\pizhalf{} / 10      & \phantom{0}86.4\%   & \phantom{0}56.6\%   & ---                 \\
\pizhalf{} / spatial & \phantom{0}91.0\%   & \phantom{0}55.0\%   & ---                 \\
OFT / object         & 100.0\%             & \phantom{0}96.0\%   & ---                 \\
OFT / goal           & \phantom{0}98.0\%   & \phantom{0}12.0\%   & ---                 \\
OFT / spatial        & \phantom{0}92.0\%   & \phantom{0}58.0\%   & ---                 \\
X-VLA / object       & 100.0\%             & \phantom{0}60.0\%   & 60.0--90.0\%        \\
X-VLA / goal         & \phantom{0}94.0\%   & \phantom{0}10.0\%   & \phantom{00}4.0--10.0\%      \\
X-VLA / spatial      & \phantom{0}98.0\%   & \phantom{0}48.0\%   & 44.0--58.0\%        \\
SmolVLA / object     & \phantom{0}89.2\%   & \phantom{00}0.0\%   & ---                 \\
SmolVLA / MW easy    & \phantom{0}85.0\%   & \phantom{0}82.0\%   & ---                 \\
SmolVLA / MW hard    & \phantom{0}62.0\%   & \phantom{0}41.0\%   & ---                 \\
GR00T / object       & \phantom{0}93.0\%   & \phantom{0}70.0\%   & \phantom{0}50.0\%   \\
GR00T / goal         & \phantom{0}97.0\%   & \phantom{00}0.0\%   & \phantom{00}0.0\%   \\
GR00T / long         & \phantom{0}83.0\%   & \phantom{0}67.0\%   & \phantom{0}47.0\%   \\
\bottomrule
\end{tabular}
\end{table}

Even though behavior is largely invariant to the prompt, a linear probe on the late layers of each language-capable model classifies the prompt category with over 99\% accuracy (Tab.~\ref{tab:cross-model-probing}). All models encode the prompt and use it only when they cannot determine the goal from the scene. \citet{huang2026delock} found similar results on physical robots and attribute them to visual-encoder drift (App.~\ref{sec:extended-related}).

\subsection{VLM Pathways Select the Goal and Expert Pathways Generate the Motor Program}
\label{sec:pathway-specialization}

The bind/select division of labor predicts an architectural division, with VLM pathways selecting the goal and motor-expert pathways generating the coordinate-bound motor program. The dissociation is quantified on SmolVLA and reproduced directionally on the other two multi-pathway models. Injecting expert pathways causes active misdirection (reaching to wrong locations), and injecting VLM pathways causes passive stalling (timeout) (Fig.~\ref{fig:displacement}, right). On SmolVLA, oracle state-prediction probes recover ground-truth state from expert activations at $4.5\times$ the VLM rate (oracle ratio 0.58 vs.\ 0.13, Tab.~\ref{tab:cross-model-probing}). The prediction has two halves, the VLM side decoding the goal better and the expert side decoding the state better. SmolVLA shows the state half and \pizhalf{} the goal half (a modest 14\,pp gap): PaliGemma goal classification rises to 76.4\% while expert goal classification settles at 62.6\% (App.~\ref{sec:goal-encoding}), and GR00T DiT layers are more ablation-sensitive (40--80\% drop) than VL-SA layers (Fig.~\ref{fig:groot-layers}), consistent with the same expert-encodes-state pattern under a different fusion. The goal side is causal. An edit to the VLM pathway at the instruction positions restores goal selection with no instruction on four of five architectures, and matched random or wrong-task edits do not. OFT, which has no separate VLM pathway, does not respond (App.~\ref{sec:silent-steering}). Where a VLA has a separate expert, the expert holds the coordinates and the VLM holds the goal and an edit on the VLM side selects it.

\subsection{Pooling Preference Is Architecture-Tied and Temporal Contrast Collapses Phase}
\label{sec:sae-analysis}

Reading this substrate with SAEs requires us to decide how we want to pool over action tokens, which can vary by model architecture. We train TopK SAEs on all pathway layers (Eqs.~\ref{eq:sae-encoder}--\ref{eq:sae-loss}; 520+ SAEs total) and observe two trends across architectures.

Single-feature ablation has no impact on 28--92\% of concept/task combinations (Fig.~\ref{fig:concept-ablation-cross-model}), consistent with concepts being distributed redundantly across multiple features. Certain layers show higher sensitivity, and the action pathway differs from the language stack (Fig.~\ref{fig:cross-arch-killswitch}). Both the \pizhalf{} expert (18 layers) and GR00T DiT (16 layers) show that 40\% or more of ablated feature/task combinations kill the policy ($\Delta$SR $\geq 0.5$) for the majority of their depth, while OFT (32 layers), X-VLA (24 layers) and the SmolVLA expert (32 layers) keep this kill-switch rate under 10\% at almost every layer. Model width does not predict feature sensitivity (see model-specific charts in App.~\ref{sec:additional-results}).

We compare the five models with a three-arm rollout setup and observe two behaviors when pooling features (Tab.~\ref{tab:cross-arch-3arm-rollout}, rollout success rate through the per-token SAE and the mean-pooled SAE). \pizhalf{} drops from 81\% to 3\% and OFT from 80\% to 7\% when using a mean-pooled SAE, indicating they both need to be reconstructed on a per-token basis. X-VLA sees an increase (51\% to 80\%) and GR00T-DiT performs slightly better (80\% to 83\%), with SmolVLA staying constant (67.2\% and 66.4\%). With the temporal-contrastive loss we also observe the collapse of adjacent codes (adjacent-code cosine similarity $\cos_{\text{adj}}$ 0.86 to 0.997) but still have an $R^2$ of 0.97 or more on \pizhalf{}, X-VLA and GR00T-DiT (SmolVLA 0.940; OFT is data-limited). On all models we tested, this collapse comes with a drop in rollout success. Pooling should be chosen per model and signal type (App.~\ref{sec:silent-steering}), and the temporal-contrastive loss should be avoided. All mean-pool values are computed with the replacement write-back. When we use a residual write-back instead, the mean-pool collapse disappears on OFT, GR00T and X-VLA (App.~\ref{sec:sae-k-sweep}).

\subsection{The Program Replicates on a Physical UR10e}
\label{sec:physical-body}

\begin{figure}[t]
    \centering
    \includegraphics[width=0.55\linewidth]{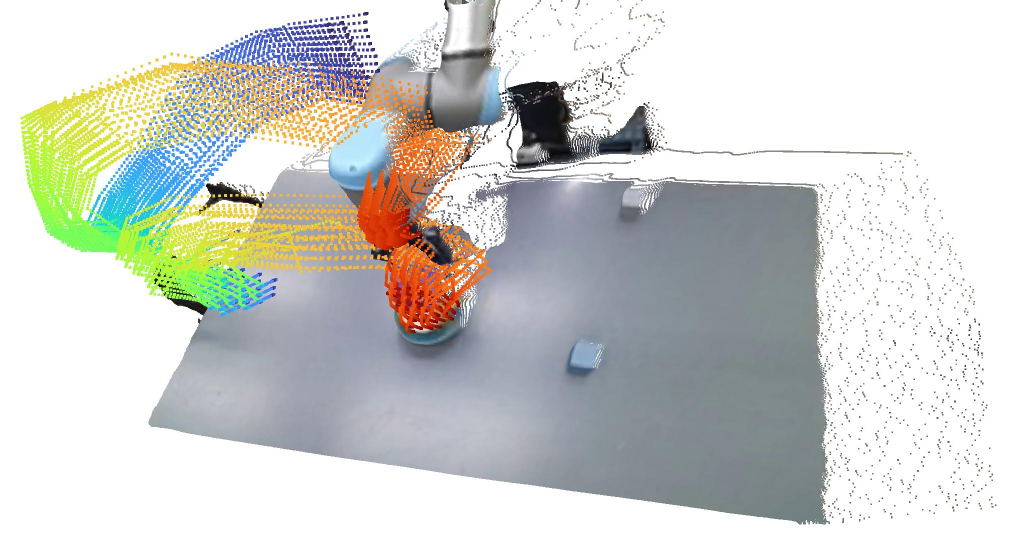}
    \caption{\textbf{Directed injection on the UR10e.} The plushie program is injected at expert L16/L17 under the prompt to stack the blocks. The arm reaches the plushie and carries it to the bowl, and the blocks stay untouched. Tracked points run from start (blue) to end (red).}
    \label{fig:physical-tracks}
\end{figure}

We take the claim from \S\ref{sec:cross-task} out of the simulator and fine-tune \pizhalf{} on a UR10e. We test 7 tabletop tasks (76.2\% base success rate, 160/210, with 30 rollouts per task) and perform the same wholesale expert replay at L16/L17 from plushie-in-bowl to gray-on-blue stacking and vice versa (App.~\ref{sec:physical}). When injecting our directed activation (Tab.~\ref{tab:physical-injection}) we see the robot perform the source task 26/26 times in one direction and 18/26 in the other. This is compared to 0/24 and 0/25 with no injection on the same scene, and 0/25 and 1/23 for a random direction with the same norm. To separate object tracking from coordinate anchoring, we then displace the source object 10--15\,cm from where it was recorded. When we inject the activation the robot goes to the recorded coordinates in 50/50 trials across both directions. The non-injected policy goes to the displaced object (Fig.~\ref{fig:physical-tracks}; object tracking done with CoTracker3~\citep{karaev2024cotracker3} and DA3~\citep{lin2025depthanything3}). We also reuse the flow-matching sampling noise from the source task when doing physical replays. This is not matched in our simulation results (App.~\ref{sec:physical}).

\begin{table}[t]
\centering
\caption{Physical cross-task injection on the UR10e, source-classified rate per condition [95\% CI].}
\label{tab:physical-injection}
\footnotesize
\begin{tabular}{@{}lcc@{}}
\toprule
\textbf{Condition} & \textbf{Blocks program under plushie prompt} & \textbf{Plushie program under blocks prompt} \\
\midrule
Directed injection      & \textbf{26/26} = 100.0\% [87.1, 100.0] & \textbf{18/26} = 69.2\% [50.0, 83.5] \\
No injection (floor)    & 0/24 = 0.0\% [0.0, 13.8]               & 0/25 = 0.0\% [0.0, 13.3] \\
Norm-matched random     & 0/25 = 0.0\% [0.0, 13.3]               & 1/23 = 4.3\% [0.8, 21.0] \\
\bottomrule
\end{tabular}
\end{table}

\subsection{Limitations}
\label{sec:limitations-body}

Simulation carries all cross-architecture results, and hardware covers a single model and task pair (\S\ref{sec:physical-body}). Extended limitations and three falsifiable predictions are in App.~\ref{sec:limitations}.

\section{Conclusion}
\label{sec:conclusion}

Across six architectures, fine-tuned VLAs learn to represent coordinate-bound motor programs that are selected by the vision pathway and executed by the expert pathway if it exists. When we inject these activations into the action expert we see 99.6\% of \pizhalf{} rollouts follow the source task's trajectory in a new scene. Sensitivity to language depends on the task and the architecture (\S\ref{sec:language-sensitivity}). The binding can explain the brittleness seen in benchmarks~\citep{zhou2025liberopro,fei2025liberoplus} and replicates on a real UR10e (\S\ref{sec:physical-body}). The goal is represented in the VLM and can be restored with a single edit without instructions on 4/5 architectures. When learning SAEs on VLAs, pooling should be chosen per architecture, and the temporal-contrastive loss hurts all architectures we tested.

\begin{ack}
This material is based upon work supported by the National Science Foundation Graduate Research Fellowship Program under Grant No.~2437833 and by National Science Foundation Grant No.~2533634. The authors also thank NVIDIA for computing resources provided through the NVIDIA Academic Grant Program.
\end{ack}

\bibliographystyle{plainnat}
\bibliography{references}

\newpage
\appendix

\begin{center}\footnotesize
\begin{tabular}{@{}lp{0.66\linewidth}l@{}}
\toprule
\textbf{App.} & \textbf{Contents} & \textbf{Supports} \\
\midrule
\ref{sec:extended-related} & VLA architectures, interpretability, robustness benchmarks & \S\ref{sec:related} \\
\ref{sec:limitations} & Limitations and falsifiable predictions & \S\ref{sec:cross-task}, \S\ref{sec:limitations-body} \\
\ref{sec:discrete-act} & Controls: discrete-token OpenVLA, ACT, random-direction injection & \S\ref{sec:cross-task} \\
\ref{sec:appendix-methodology} & SAE training, concept scoring, protocols, architectures, benchmarks, compute & \S\ref{sec:method} \\
\ref{sec:cross-arch-synthesis} & Cross-model radar and the per-architecture pathway metrics & \S\ref{sec:pathway-specialization} \\
\ref{sec:component-level} & Goal-encoding dimensions, FFN neurons, layer similarity on \pizhalf{} & \S\ref{sec:pathway-specialization} \\
\ref{sec:silent-steering} & Goal steering with no instruction, on five architectures & \S\ref{sec:pathway-specialization} \\
\ref{sec:additional-results} & Tie-inclusive override, steering sensitivity, SAE pooling arms, DTW & \S\ref{sec:visual-dominance}--\S\ref{sec:sae-analysis} \\
\ref{sec:ablation-studies} & GR00T phase-conditioned feature ablation & \S\ref{sec:sae-analysis} \\
\ref{sec:qualitative-results} & Qualitative rollouts and additional figures & all \\
\ref{sec:per-arch-deepdive} & Per-architecture failure modes, OFT per suite, cross-model probing & \S\ref{sec:visual-dominance}--\S\ref{sec:language-sensitivity} \\
\ref{sec:dose-response} & Dose-response of expert injection & \S\ref{sec:cross-task} \\
\ref{sec:sae-k-sweep} & SAE sparsity sweep and residual write-back & \S\ref{sec:sae-analysis} \\
\ref{sec:datascale} & Override under a tenfold cut in fine-tuning data & \S\ref{sec:cross-task} \\
\ref{sec:physical} & UR10e setup, directed injection, displaced-object test & \S\ref{sec:physical-body} \\
\ref{sec:broader-impact} & Broader impact &  \\
\bottomrule
\end{tabular}
\end{center}

\section{Extended Related Work}
\label{sec:extended-related}

\subsection{Vision-Language-Action Models}

RT-2~\citep{brohan2023rt2} showed that web-pretrained VLMs can be fine-tuned for manipulation via tokenized actions. OpenVLA~\citep{kim2024openvla} extended this with an open-source 7B model, and OpenVLA-OFT~\citep{kim2025openvlaoft} replaced discrete tokenization with continuous L1 regression, achieving 97.1\% LIBERO success. The $\pi_0$, \pizhalf{}, $\pi^*_{0.6}$, and $\pi_{0.7}$ models~\citep{black2024pi0,pi05,physicalintelligence2025pistar06,physicalintelligence2026pi07} introduced flow matching with a dedicated action expert and progressively more diverse training context. Whether $\pi_{0.7}$'s reported emergent capabilities reflect compositional generalization or further-scaled memorized motor patterns is an unexplored question that these mechanistic tools could possibly explain. Current VLAs span several scales and paradigms: SmolVLA~\citep{shukor2025smolvla} (450M), X-VLA~\citep{zheng2025xvla} (cross-embodiment), GR00T N1.5~\citep{bjorck2025groot} (humanoid), CogACT~\citep{cogact}, GR-3~\citep{gr3}, VITA-VLA~\citep{vitavla}, SimpleVLA-RL~\citep{li2025simplevlarl}, RIPT-VLA~\citep{riptvla}, FLOWER~\citep{flower}, SpatialVLA~\citep{qu2025spatialvla}, and ACT~\citep{zhao2023learning} (80M, no language). Modular approaches include SayCan~\citep{ahn2022saycan}, PaLM-E~\citep{driess2023palme}, and Code as Policies~\citep{liang2023code}. Concurrent representation-analysis work in robot policies uses different tools: \citet{lei2026simreal} apply UMAP and Wasserstein-distance analysis to characterize sim-real co-training, identifying ``structured representation alignment'' and ``importance reweighting'' effects. Our SAE-based feature-level analysis is complementary.

\paragraph{Action Representation.}
Diffusion Policy~\citep{chi2023diffusion} treats action generation as conditional denoising, 3D Diffusion Policy~\citep{ze2024dp3} incorporates 3D representations, and FAST~\citep{pertsch2025fast} uses DCT-based compression tokenization. Action tokenization directly determines SAE applicability (Section~\ref{sec:discrete-vla}).

\paragraph{Language Following.}
CAST~\citep{zhang2025cast} augments robot datasets with counterfactual language labels. VLM4VLA~\citep{vlm4vla} found that VLM benchmark performance does not predict VLA task success, so language understanding and visuomotor competence are decoupled.

\subsection{Mechanistic Interpretability}

SAEs~\citep{olshausen1997sparse,bricken2023monosemanticity,cunningham2023sparse,gao2024scaling,templeton2024scaling} decompose dense activations into sparse interpretable features, with extensions to vision~\citep{rao2025sparse,clipsae1,cliphirearchy} and vision-language models~\citep{lim2025sparse}. Activation steering~\citep{turner2023activation,rimsky2024steering} enables behavioral control, with advances including conceptor-based steering~\citep{wang2024conceptors}, SAE-guided additions~\citep{wu2025fgaa}, and VLM safety steering~\citep{chen2025spovlm,li2025steervlm}. Linear probing~\citep{alain2017understanding,belinkov2022probing} identifies linearly decodable information, and establishing causal relevance requires intervention~\citep{wang2023interpretability}. PvP~\citep{zhang2024pvp} shows vision and priors compete in multimodal models.

\subsection{Interpretability for Robot Learning}

RT-1~\citep{brohan2022rt1} included attention visualizations, Diffusion Policy~\citep{chi2023diffusion} analyzed action distributions, and ALOHA~\citep{zhao2023learning,aloha2team2024} demonstrated bimanual manipulation, all without internal representation analysis. RoboFail~\citep{robofail2024}, AHA~\citep{duan2024aha}, and RACER~\citep{dai2025racer} characterize failures behaviorally. \citet{haon2025mech} introduced VLA steering via internal representations, \citet{molinari2025worldmodel} probed for emergent world models, and \citet{khan2025sparse} used SAEs to isolate interpretable steering directions in Magma. Interpretability-by-design approaches include CoA-VLA~\citep{li2025coavla}, CoT-VLA~\citep{zhao2025cotvla}, TriVLA~\citep{trivla}, dVLA~\citep{dvla}, and Embodied-R1~\citep{embodiedr1}.

\paragraph{Cross-Model and Multimodal SAE Alignment.}
Training SAEs on individual models, ours included, leaves each model with its own feature space. This makes it difficult to compare similar concepts across different pathways or models. There have been two directions of work that attempt to solve this. One is to scale SAEs across models. Universal Sparse Autoencoders~\citep{thasarathan2025usae} and SPARC~\citep{sparc2025} train a common dictionary across models. They force concepts to be aligned by sharing an encoder or using a global TopK constraint. Group-SAE~\citep{groupsae2024} shares a dictionary between nearby layers based on similarity in the residual stream to reduce compute. Another direction is to align modalities. \citet{groupsparse2026multimodal} use a cross-modal random mask and a group-sparse penalty to mitigate the split dictionary effect, where features fire for only one modality. This is similar to the VLM/expert pathway split we find (\S\ref{sec:pathway-specialization}), and if we trained an SAE that was aligned across modalities we could test whether the goal/motor split is orthogonal at the dictionary level or is an artifact of training each SAE separately. We leave this as future work.

\paragraph{Concurrent VLA Architectures.}
There are many VLA architectures with different methods of decoding actions that we can use our methods to study. AR-VLA~\citep{arvla2026} autoregressively generates actions with a longer memory of previous actions, allowing it to generate actions causally instead of in chunks. DFM-VLA~\citep{dfmvla2026} uses discrete flow matching to generate actions iteratively instead of a static set of tokens. AsyncVLA~\citep{asyncvla2025} uses asynchronous flow matching and has a rater that decides when to regenerate actions. HybridVLA~\citep{hybridvla2025} combines both autoregressive and diffusion-based decoding in one LLM. These all change how the model will generate an action given a scene. However, our method of injecting tasks and zeroing out pathways can be applied to these architectures. We discuss our issue with where to place the hook in App.~\ref{sec:hook-protocol}. It would be interesting to see if our findings about coordinate binding hold when actions are refined iteratively, regenerated based on confidence, or generated with more context from previous actions (DFM-VLA, AsyncVLA, AR-VLA); we leave this to future work. We believe that the more an architecture re-grounds on the current observation between action chunks, with no commitment to a precomputed plan, the weaker the binding will be.

\paragraph{Single-Policy Diagnostics.} VLA-Trace~\citep{shi2026vlatrace} traces \pizhalf{} and OpenVLA across checkpoints using kernel alignment and attention knockout. They find that the models are able to ground in the visual scene but struggle with following instructions in detail, which aligns with our findings in \S\ref{sec:language-sensitivity}. LAVLA~\citep{wulff2026lavla} clusters the action decoder latents in GR00T N1.5 layer by layer and finds spatio-temporal and kinematic structure separating in the middle layers. Neither of these studies looks at more than 2 models, and neither intervenes on the task. \citet{huang2026delock} report the same collapse of a fine-tuned \pizhalf{} on hardware when objects move. They believe this is due to drift in the visual encoder. Our probes find grounding intact (over 99\% decodable), showing that the issue lies later, in the action pathway.

\section{Limitations}
\label{sec:limitations}

\paragraph{Predictions.} If the expert is learning a motor program that is bound to coordinates and selected by the scene, three predictions follow. (i) Moving objects out of their training-distribution positions should cause the model to fail while keeping the instruction the same. Robustness benchmarks have shown that models struggle with position changes~\citep{zhou2025liberopro,fei2025liberoplus}. (ii) For semantic control, one would need to change the pathway that selects the goal. Our pathway-injection results are consistent with this, since injecting into the expert causes the model to go to the wrong location and injecting into the VLM causes it to not move. \citet{jin2026event} were unable to semantically steer the model on the feed-forward side, which we would also expect. (iii) As fine-tuning data becomes more diverse, the ability to override the model should decrease. \citet{swann2026sparse} show this at the feature level. But the number of demonstrations alone does not affect the ability to override (see App.~\ref{sec:datascale} for no change in override across a 10$\times$ range of demonstrations).

\subsubsection*{Benchmark Scope}
All cross-architecture results come from four simulated benchmarks. The UR10e covers \pizhalf{} on one task pair (App.~\ref{sec:physical}) and tests injection and coordinate binding under real sensing and actuation, while visual dominance and pathway specialization remain simulation results. Fine-tuning on a new robot reshapes internal representations, so the next step is applying the same protocol to other architectures on additional embodiments. We leave this to future work, but invite the contribution to Action Atlas.

\subsubsection*{Coordinate-Frame Disambiguation}
Our injection and displacement experiments (App.~\ref{sec:dtw-cross-task}) fit binding to absolute workspace coordinates. It could also be that the model is bound to coordinates relative to the scene that happen to line up with our source task. If we were to apply a rigid transform to the entire world in our simulation we could differentiate between these two cases, but we did not try this. Either way the model is choosing a path that is bound to coordinates and does not re-ground on the visible task. The physical displaced-object test (\S\ref{sec:physical-body}) separates object tracking from coordinate anchoring.

\subsubsection*{Counterfactual Prompt Coverage}
The prompts that we use to test if the model is ignoring language are very basic (negation, null, swap). Stronger evidence for our language-insensitivity claims would come from more complex instructions that require the model to understand multiple objects and actions, such as ``Pick up the red cup and put it behind the blue bowl'', along with ambiguous multi-object scenes. There are also benchmarks such as LIBERO-PRO~\citep{zhou2025liberopro} that test models under corrupted instructions and changes to the environment, a stricter test of language grounding.

\subsubsection*{Cross-Task Injection Confounds}
When injecting into the model we are creating off-manifold activations, and if we are not careful with timing we could put the model in an invalid state. In App.~\ref{sec:random-injection} we list three observations that weigh against a drift explanation. First, if we inject a random direction with the same norm as our injected activation, it is at least as far from the manifold and gives no better than chance override. Second, displacement analysis places the trajectories closer to the source task than to the destination task. Third, on hardware it is the random arm that moves incoherently (App.~\ref{sec:physical}).

\subsubsection*{Concept Identification}
SAE features are not guaranteed to be disentangled. Ablation validation finds that some identified features encode general motor primitives and lack a clean semantic reading.

\section{Methodology Controls}
\label{sec:discrete-act}

\subsection{Discrete Tokenization Prevents SAE Intervention}
\label{sec:discrete-vla}

Before adopting OpenVLA-OFT, we conducted experiments on base OpenVLA (autoregressive discrete 256-bin tokenization). Despite successful SAE training ($R^2 = 0.87$--$0.96$), hooking SAEs into the forward pass produced 0\% task success on all but the final layer. The discrete tokenization maps activations to bins via argmax. Even small reconstruction errors shift the selected bin, with errors compounding across the 7-token autoregressive sequence. Replacing discrete tokenization with continuous L1 regression via OpenVLA-OFT enables SAE intervention at 99.2\% success. Scale is fixed at 7B, so the action representation sets SAE applicability.

\subsection{ACT as Non-VLM Control}
\label{sec:act-control}

ACT is a control that does not have a language pathway to act through. There is no effect at this hook when doing cross-task injection between TransferCube and Insertion (1.0 cosine sim and all bits are the same in the action array). So ACT cannot demonstrate cross-task evidence. Grid ablations reveal spatially structured representations. We found that if we ablate the grid at index (2,2) where the majority of the task is completed we see a drop in success on both tasks (Tab.~\ref{tab:null-injection}). Adding Gaussian noise with a std of 0.1 makes the TransferCube task fail 100\% of the time and the Insertion task 75\%, which shows that visual-pathway dominance holds for a vision-only policy.

\subsection{Random-Direction Injection Control}
\label{sec:random-injection}

An argument can be made that perturbing the residual stream enough will cause the policy to perform whatever in-distribution motor behavior it is closest to. In order to test this we used \pizhalf{}, \texttt{libero\_goal} and expert layers L16+L17 (the same hooks as the main cross-task condition). During each forward pass $t$ we changed the injection $\mathbf{h}^A_\ell$ to $r_t \cdot \|\mathbf{h}^A_\ell\|_2$, a random Gaussian direction $r_t \sim \mathcal{N}(0, I)$ normalized to the unit sphere and scaled by the norm of the injection from the source task. Destination scene and source prompt remained the same. In App.~\ref{sec:dose-response} we vary how strongly we inject the directed activations.

Over the course of 50 episodes (10 pairs $\times$ 5 seeds) we were able to observe 22\% source behavior, 42\% destination behavior, 36\% ambiguous and a success rate of 0\%. Cosine similarity to source task was on average $-0.017$ and 0.012 to destination task. These numbers are roughly equal to 0. When we calculated our override rate (did not use a threshold, just compared cosine similarity to source vs dest) we achieved 50\% [36.6--63.4], which is comparable to chance.

When performing our cross-task injection we achieved a 99.6\% source behavior rate using the same test suite and layers (1,614 episodes), a difference of 78pp with no overlap in confidence intervals. This shows that the injected direction is what allows for overriding. Injecting a random direction with the same norm only throws the policy off.

\paragraph{Replication at $n{=}200$.} With the destination prompt held on its own task text (10 pairs $\times$ 10 episodes per suite, with seeds starting at 42), we see a 48.0\% override rate on \texttt{libero\_goal} and 43.0\% on \texttt{libero\_object}. Pooled together this is 45.5\% (200 episodes), consistent with the 50\% chance level. Mean cosine similarity to the source trajectory was 0.01 and to the destination trajectory was 0.02 on both suites. Success rate on the destination task was 30\% on goal and 0\% on object, compared to 95\% and 92\% respectively with no injection. So we can throw off the policy with a random direction but not cause it to perform the source task.

\paragraph{Off-manifold distance does not predict override.} Injected source activations are off-distribution by construction. We take a Gaussian fit of the expert layers on 6 \texttt{libero\_goal} tasks (8k samples per layer) and score our held-out natural samples against our injected samples from the other 4 tasks. We get a Mahalanobis distance of 46.9 on injected samples vs 18.0 on natural samples at L16 (median distances of 40.7 and 17.5). Only 0.4\% of injected samples fall within the 95th percentile of natural samples. At L17 we get a distance of 34.4 vs 16.7 with 6.5\% of injected samples falling within the 95th percentile. If we include the source tasks in our distribution we see the gap decrease (L16 32.9 vs 20.0 with 9.1\% within the 95th percentile; L17 24.8 vs 17.9 with 38.9\% within the 95th percentile). Our random direction is at least as far away from the natural distribution and still gives chance-level override. So being far away from the natural distribution is not what is causing our override. If the policy were just drifting off manifold we would expect to see incoherent movement. When we inject the source direction we see the policy produce a complete source-task trajectory and move towards the coordinates of the source task. The random direction on the UR10e is what causes the policy to move incoherently (App.~\ref{sec:physical}).

\section{Extended Methodology Details}
\label{sec:appendix-methodology}

\subsection{SAE Architecture and Training}
\label{sec:sae-architecture}

The 520+ SAEs of Tab.~\ref{tab:episode-counts} are \pizhalf{} 72 (18 layers $\times$ 2 pathways $\times$ 2 pooling), OFT 96 (32 layers $\times$ 3 arms), X-VLA 120 (24 layers $\times$ 2 pooling $\times$ 2 environments, plus temporal), SmolVLA 146 (32 layers $\times$ 2 pathways $\times$ 2 pooling, plus 18 temporal), and GR00T 87 (32 layers $\times$ 2 pooling, plus 23 temporal). Concepts total 79 unique names, or 82 if the three names shared by the spatial and object categories count once per category.

\paragraph{Architecture.} Following the TopK design of \citet{gao2024scaling}, our sparse autoencoders consist of an encoder-decoder pair with tied weights. For an input activation $\mathbf{h}\in\mathbb{R}^{d}$ (per token; $d \in \{480, 960, 1024, 1536, 2048, 4096\}$ across the five VLAs),
\begin{align}
\mathbf{z} &= \mathrm{TopK}_{k}\!\left(\mathbf{W}_e (\mathbf{h} - \mathbf{b}_d) \right), \label{eq:sae-encoder}\\
\hat{\mathbf{h}} &= \mathbf{W}_e^\top \mathbf{z} + \mathbf{b}_d, \label{eq:sae-decoder}
\end{align}
where $\mathbf{W}_e \in \mathbb{R}^{m \times d}$ with expansion factor $m \in \{4d, 8d\}$, and $\mathrm{TopK}_k(\mathbf{u})$ retains the $k=64$ largest pre-activations and zeros the remainder. Tied decoder weights ($\mathbf{W}_d=\mathbf{W}_e^\top$) and the pre-encoder bias subtraction follow~\citet{bricken2023monosemanticity, gao2024scaling}.

\paragraph{Training objective.} TopK allows us to enforce sparsity without needing to add an L1 penalty~\citep{cunningham2023sparse}. We optimize reconstruction MSE plus an AuxK term~\citep{gao2024scaling}. AuxK recycles dead features by reconstructing the residual $\mathbf{r} = \mathbf{h} - \hat{\mathbf{h}}$ (input $-$ reconstruction) using the top $k_{\text{aux}}{=}512$ inactive features:
\begin{equation}
\mathcal{L}(\mathbf{h}) \;=\; \underbrace{\|\mathbf{h} - \hat{\mathbf{h}}\|_2^2}_{\text{reconstruction}} \;+\; \alpha_{\text{aux}} \, \underbrace{\|\mathbf{r} - \mathbf{W}_e^\top \mathrm{TopK}_{k_{\text{aux}}}(\mathbf{z}_{\text{dead}})\|_2^2}_{\text{AuxK (dead-feature recycling)}},
\label{eq:sae-loss}
\end{equation}
with $\alpha_{\text{aux}} = 1/32$. Decoder columns are unit-normalized after each step.

\paragraph{Training Hyperparameters.}
Each SAE was trained on 500k activations ($\sim$10k forward passes $\times$ 50 action tokens) with a batch size of 4096 for 100 epochs. The learning rate was $3 \times 10^{-4}$ with cosine decay. We used TopK to select $k{=}64$ active features per token. Decoder weights were tied to the encoder transpose ($\mathbf{W}_d = \mathbf{W}_e^\top$).

\subsection{Concept-Based Feature Identification}
\label{sec:concept-identification}

Features associated with concepts were selected using the frequency-weighted contrastive score described in \S\ref{sec:sae} ($\text{score}_f = d_f \times \text{freq}_f$). The active set is the top 64.

With TopK sparsity, features with high mean activation across samples do not necessarily appear in the active top-64 for individual samples, reducing their causal relevance.

\paragraph{Sample population for Cohen's $d$.} Activations were max-pooled over the 50 action tokens in each episode, giving us an activation vector for each (episode, layer, feature). Cohen's $d_f$ was calculated over episodes in each concept set ($n{=}100$--$200$ episodes per concept). We take the difference in mean between episodes that have the concept and those that do not and divide by the std. We use unbiased Welch variance for our population stats.

\paragraph{Concept specificity scores in Tab.~\ref{tab:pi05-concept-features}(a).} The dimensionless score $\text{score}_f = d_f \times \text{freq}_f$ is bounded above by $|d_f|$ and is at most a few units of effect size in our data. The values reported in parentheses in Tab.~\ref{tab:pi05-concept-features}(a) (e.g., 133k, 412k) are raw cumulative activation magnitudes per concept, computed as $\sum_{(\text{ep}, t) \in \mathcal{C}} \mathrm{ReLU}(z_{f,t,\text{ep}})$ over the concept-positive episode set $\mathcal{C}$ and the 50 action-token positions, summed across layers. We retain the raw-magnitude column for readability of layer rankings within a concept. The per-feature dimensionless $\text{score}_f$ used to identify concept-associated features is reported in the supplementary feature-exploration platform.

\subsection{Ablation Protocol}
\label{sec:ablation-protocol}

Feature ablation is performed by zeroing selected features in the SAE latent space:
\begin{align}
\mathbf{z}_{\text{ablated}} &= \mathbf{z} \odot \mathbf{b} \\
\mathbf{h}_{\text{modified}} &= \mathbf{h} + \mathbf{W}_e^\top (\mathbf{z}_{\text{ablated}} - \mathbf{z})
\label{eq:sae-ablation}
\end{align}
where $\mathbf{b} \in \{0,1\}^{m}$ is a binary mask (with $m$ the SAE expansion dimension) carrying zeros at ablated feature indices. Per-token ablation applies this independently to each of the 50 action tokens. Writing the modification as a residual ($\mathbf{W}_e^\top(\mathbf{z}_{\text{ablated}}-\mathbf{z})$) preserves the unmodeled component of the activation and intervenes only on what the SAE explains.

\paragraph{Steering operator.} The ablation operator generalizes to a continuous steering operator that modulates a target set of features $\mathcal{F}$ by a scalar coefficient $\alpha$:
\begin{equation}
\mathbf{z}_{\text{steered},f} \;=\;
\begin{cases}
\alpha\, \mathbf{z}_f & f \in \mathcal{F},\\
\mathbf{z}_f & \text{otherwise},
\end{cases}
\qquad
\mathbf{h}_{\text{steered}} \;=\; \mathbf{h} + \mathbf{W}_e^\top(\mathbf{z}_{\text{steered}} - \mathbf{z}).
\label{eq:sae-steering}
\end{equation}
$\alpha = 0$ recovers ablation, $\alpha = 1$ is identity, $\alpha \in (0,1)$ dampens, $\alpha < 0$ negates, and $\alpha > 1$ amplifies. Because TopK enforces hard sparsity, $\alpha\,\mathbf{z}_f = 0$ whenever feature $f$ is inactive in the top-$k$ set. This multiplicative operator therefore modulates only currently-active concepts and cannot inject features absent from the active set. Additive steering ($\mathbf{z}_f \leftarrow \mathbf{z}_f + \alpha\,\mathbf{v}_f$ for a steering vector $\mathbf{v}_f$) injects concepts outside the active set, and App.~\ref{sec:silent-steering} applies it in the VLM pathway. The cross-architecture sensitivity Fig.~\ref{fig:concept-ablation-cross-model} reports ablation ($\alpha=0$).

\subsection{Intervention-Hook Protocol}
\label{sec:hook-protocol}

Activation interventions hook one of two locations: the full \emph{layer output} (residual stream after attention $+$ MLP $+$ residual addition) or, where the layer wrapper is bypassed, the \emph{MLP output} that writes into the residual stream. Some VLAs bypass the standard DecoderLayer in their inference path, so hooking the layer wrapper does not modify what the policy consumes.

\paragraph{Per-architecture hook placement.}
All interventions hook the layer output (i.e., full residual stream): cross-task injection (Eq.~\eqref{eq:injection}), layer zeroing, SAE training, SAE reconstruction and ablating/steering SAE features (Eqs.~\eqref{eq:sae-ablation} and~\eqref{eq:sae-steering}), meaning each SAE will act on the activations it was trained on. SmolVLA is an exception because its interleaved VLM-expert loop invokes the submodules of each layer directly and does not call the wrapper \texttt{DecoderLayer.forward}. So we hook \texttt{layer.mlp.forward}, which is the MLP writing to the residual stream, for all these interventions. Goal-steering interventions from App.~\ref{sec:silent-steering} directly affect the residual stream of SmolVLA by reading out the post-attention residual using a pre-hook on the layer's post-attention norm, adding the MLP output and writing out using the MLP output. We confirmed hook visibility using cosine identity ($\cos{=}1.0$ between hooked and unhooked baselines when doing no-op interventions).

\paragraph{Reading the full-layer interventions.}
The OFT 14--15\% recovery under null prompt plus baseline injection in Tab.~\ref{tab:null-injection}, and the X-VLA and GR00T 0\% under single-layer zeroing, replace the full layer output, so they are upper bounds on layer importance under residual-stream patching. On SmolVLA the same interventions change only the MLP write and leave the residual passthrough intact.

\begin{table}[t]
\centering
\caption{Architectures under study. PG = PaliGemma, Exp.\ = Expert, Flor.\ = Florence, D+E+V = DiT + Eagle + VL-SA, LIB.\ = LIBERO, MW = MetaWorld, SimpE.\ = SimplerEnv. Details in App.~\ref{sec:model-architecture}.}
\label{tab:architectures}
\resizebox{\columnwidth}{!}{%
\begin{tabular}{@{}lcccccc@{}}
\toprule
\textbf{Model} & \textbf{Params} & \textbf{Layers} & \textbf{Dim} & \textbf{Action Gen.} & \textbf{Pathway} & \textbf{Bench.} \\
\midrule
\pizhalf{}~\citep{pi05} & 3B & 18 & 1024 & Flow (10 steps) & Dual (PG + Exp.) & LIBERO, UR10e \\
OFT~\citep{kim2025openvlaoft} & 7B & 32 & 4096 & Cont.\ L1 regr. & Single (Llama-2) & LIBERO \\
X-VLA~\citep{zheng2025xvla} & $\sim$1B & 24 & 1024 & Flow matching & Single (Flor.-2) & LIB., SimpE. \\
SmolVLA~\citep{shukor2025smolvla} & 450M & 32 & 960/480 & Flow matching & Dual (VLM + Exp.) & LIB., MW \\
GR00T~\citep{bjorck2025groot} & 3B & 32 & varies & Diff./flow & Triple (D+E+V) & LIBERO \\
ACT~\citep{zhao2023learning} & 80M & -- & -- & CVAE & Enc.-Dec. & ALOHA \\
\bottomrule
\end{tabular}}
\end{table}

\begin{table}[!htbp]
\centering
\caption{We performed experiments at scale on 6 models. Number of episodes is the number of rollouts summed over every experiment type the model participated in. Number of SAEs includes all per-token, mean-pool and temporal arms (see App.~\ref{sec:sae-architecture}). There is some overlap between models when counting concepts, and the total is the union of all concept names. $^{\ddagger}$ We only kept aggregate statistics for large ablation sweeps. Our released dataset contains per-episode data for 394{,}000+ episodes.}
\label{tab:episode-counts}
\footnotesize
\begin{tabular}{@{}lccc@{}}
\toprule
\textbf{Model} & \textbf{Episodes} & \textbf{SAEs} & \textbf{Concepts} \\
\midrule
\pizhalf{} & 71,500+ & 72 & 46 \\
OpenVLA-OFT & 70,700+$^{\ddagger}$ & 96 & 45 \\
X-VLA & 74,500+ & 120 & 55 \\
SmolVLA & 58,000+ & 146 & 60 \\
GR00T N1.5 & 164,700+ & 87 & 39 \\
ACT & 1,870 & -- & -- \\
\midrule
\textbf{Total} & \textbf{$>$420,000} & \textbf{520+} & \textbf{79} \\
\bottomrule
\end{tabular}
\end{table}

\subsection{Model Architecture}
\label{sec:model-architecture}

\subsubsection*{Pi0.5 Architecture}
\pizhalf{} uses PaliGemma (3B parameters) as its vision-language backbone and an 18-layer Gemma transformer (1024 hidden dimension) as the action expert. The action space is 7-dimensional (dx, dy, dz, dax, day, daz, gripper). The model generates 50 actions per forward pass through flow matching with iterative denoising.

\subsubsection*{OpenVLA-OFT Architecture (Continuous)}
OpenVLA-OFT is similar to OpenVLA in that it uses the same Llama-2 7B backbone and Prismatic vision encoder (DINOv2 and SigLIP). It differs by using continuous action regression instead of discrete tokens. It has an MLPResNet action head, trained with L1 loss, that predicts all 7 dimensions of the action, and it outputs 8-step action chunks (56 action tokens) in one forward pass. Optimized Fine-Tuning (OFT) uses LoRA to fine-tune the model while preserving the pretrained representation geometry.

\subsubsection*{X-VLA Architecture}
X-VLA (0.9B parameters) encodes the main camera view and the instruction with Florence-Large and generates actions with a 24-layer transformer (1024 hidden dimension) trained by flow matching. Learnable soft prompts, one set per data source, condition the shared transformer on the embodiment, and domain-specific output projections map its tokens to action chunks.

\subsubsection*{SmolVLA Architecture}
SmolVLA (450M parameters) pairs a SmolVLM2 vision-language model (960 hidden dimension) with a flow-matching action expert at half that width (480). The LIBERO checkpoint we use keeps all 32 VLM layers and a 32-layer expert. The two run layer by layer: expert layers alternate cross-attention to the VLM's keys and values with causal self-attention over the action tokens. The expert generates 50-action chunks with 10 denoising steps.

\subsubsection*{GR00T N1.5 Architecture}
GR00T N1.5 (3B parameters) has three pathways. A frozen Eagle vision-language model supplies the output of its 12th LLM layer, 4 vision-language self-attention (VL-SA) layers process those tokens, and a 16-layer diffusion transformer (DiT) generates actions. DiT blocks alternate cross-attention to the vision-language tokens with self-attention over state and action tokens, and produce 16-step action chunks with 4 flow-matching steps.

\subsubsection*{ACT Architecture}
ACT uses a ResNet-18 vision encoder (pretrained on ImageNet) feeding into a Transformer Encoder-Decoder with a CVAE latent space ($\beta = 10$, latent dimension 32). The encoder processes multi-view camera observations, and the decoder generates action chunks of 100 timesteps in 14-DOF joint space (7 per arm for bimanual manipulation).

\subsection{Benchmarks}
\label{sec:benchmark-details}

LIBERO~\citep{liu2023libero} comprises four suites of 10 tasks each: \textbf{Goal} (different goals in a shared scene), \textbf{Object} (pick-and-place with varied objects), \textbf{Spatial} (spatial reasoning and relational placement), and \textbf{LIBERO-10} (diverse tasks spanning all three categories). MetaWorld~\citep{yu2020metaworld} provides 50 tabletop manipulation tasks grouped by difficulty (easy, medium, hard, very hard). SimplerEnv~\citep{simpler_env} evaluates sim-to-real transfer on Google Robot and WidowX embodiments. ALOHA~\citep{zhao2023learning} tests bimanual manipulation with the ACT policy.

\subsection{Compute Requirements}
\label{sec:compute}

Experiments were conducted on an 8$\times$A100-SXM4-80GB cluster for \pizhalf{}, OpenVLA-OFT, large-scale SAE training, concept ablation, and cross-model analysis, an NVIDIA RTX 5090 (32GB) for GR00T N1.5 experiments, and two NVIDIA RTX 4090s (24GB) for SmolVLA and X-VLA experiments. Total experiment data exceeds 7.1 TB, including 4.3 TB of activation recordings, 152 GB of SAE checkpoints, and 420,000+ rollout episodes. The full study used approximately 10,000 A100 GPU-hours.

\subsection{Evaluation Protocol}
\label{sec:evaluation-protocol}

Episode counts per condition are stated with each experiment (20 to 50 per task for the main results, 5 per task for the earliest sweeps, and 3 per task for the OFT goal-steering screen). Episodes run for a maximum of 300 steps.

\section{Cross-Architecture Synthesis}
\label{sec:cross-arch-synthesis}

\begin{figure}[!htbp]
    \centering
    \includegraphics[width=0.5\columnwidth]{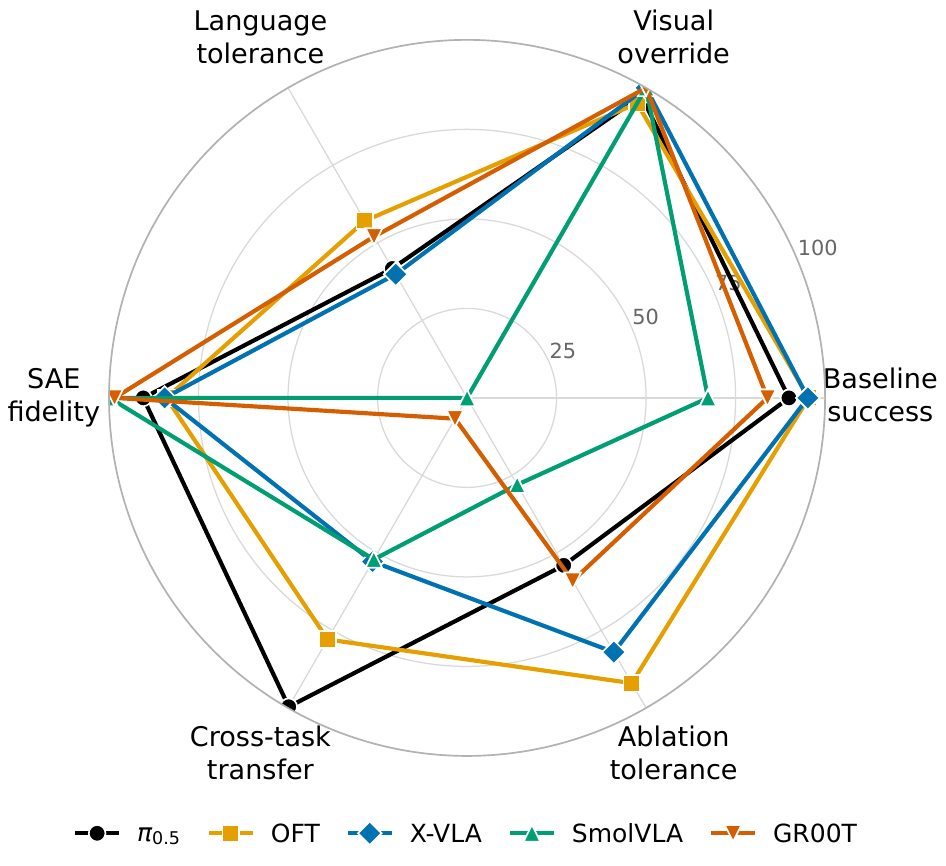}
    \caption{\textbf{Cross-model summary.} Capability radar across the five language-capable VLAs. Each axis maps a table value linearly onto 0--100, with axes defined in App.~\ref{sec:cross-arch-synthesis}.}
    \label{fig:radar-summary}
\end{figure}

Figure~\ref{fig:radar-summary} condenses the cross-architecture results of \S\ref{sec:visual-dominance}--\S\ref{sec:sae-analysis} into one view across six axes:
\begin{itemize}
    \item \textbf{Baseline success:} the suite mean of the baseline rows in Tab.~\ref{tab:null-injection}.
    \item \textbf{Visual override:} destination success lost under cross-task injection, $100 - {}$Dst.\ SR in Tab.~\ref{tab:cross-task}.
    \item \textbf{Language tolerance:} null-prompt over correct-prompt success, averaged over the LIBERO rows of Tab.~\ref{tab:counterfactual} (100 means policy ignores prompt).
    \item \textbf{SAE fidelity:} best-arm rollout success as a fraction of baseline (Tab.~\ref{tab:cross-arch-3arm-rollout}).
    \item \textbf{Cross-task transfer:} the tie-inclusive override rate (Tab.~\ref{tab:tie-inclusive}). SmolVLA's value comes from Tab.~\ref{tab:cross-task}, since its tie fraction is 0\%.
    \item \textbf{Ablation tolerance:} the zero-effect rate of single-feature ablation: \pizhalf{} 54\%, OFT 92\%, X-VLA 82\%, SmolVLA 28\%, GR00T 59\% (Fig.~\ref{fig:concept-ablation-cross-model}).
\end{itemize}
Per-architecture detail is in App.~\ref{sec:per-arch-deepdive}, and the SAE pooling results are in App.~\ref{sec:additional-results}.

The pathway-specialization claim of \S\ref{sec:pathway-specialization} rests on three architecture-specific metrics, since no single metric applies to all three multi-pathway models.

\section{Component-Level Analysis}
\label{sec:component-level}

This section traces information flow below the layer level: individual activation dimensions, FFN neurons, and layer-to-layer representational similarity. We analyze \pizhalf{} in depth (FFN weight projection and LDA on the 1024-dim expert space), then validate across OFT (32-layer neuron-level concept mapping, 80 concepts $\times$ 11,008 neurons per layer), SmolVLA (FFN neuron contrastive identification on both expert and VLM pathways), and GR00T (per-layer-type SAE feature utilization across DiT, Eagle, and VL-SA).

\subsection{Goal Encoding Dimensions}
\label{sec:goal-encoding}

Linear discriminant analysis (LDA) on \pizhalf{} expert activations at layer~17 identifies three components that separate four goal-suite tasks with 71.9\% cross-validated accuracy ($n{=}114$, 5-fold). The first component captures 45.8\% of discriminant variance, the second 35.2\%, and the third 19.0\%. The top 20 discriminant dimensions (417, 909, 934, 649, 147, 219, 708, 297, 155, 545, \ldots) define a low-dimensional subspace sufficient for goal identification within a 1024-dimensional activation space.

Goal information emerges at different rates in different pathways. For PaliGemma we see the ability to classify goals increase from 56.4\% at layer 0 to 76.4\% at layer 13 and then drop off to 68.9\% at layer 17. The expert shows a flatter trend, at 64.0\% for layer 0, with a peak of 66.4\% at layer 9 and ending at 62.6\% for layer 17. PaliGemma's steeper rise is consistent with its role as the goal encoder. The expert seems to have some knowledge about the goal throughout but never refines it.

Individual activation dimensions participate in both goal encoding and action generation. Dimension~62 most strongly modulates the x-component of action output (score 0.074), dimension~618 modulates y (0.071), and dimension~14 modulates z (0.066). The overlap between goal-discriminant dimensions (417, 909, 934) and action-modulating dimensions (62, 618, 14) is minimal, consistent with separable subspaces for goal identity and motor execution (tested below).

\paragraph{Subspace separability via subspace injection.}
We test separability by injecting only the goal-discriminant subspace (20 of 1024 dimensions from LDA) from task~A into task~B's forward pass at layer~17. Across five task pairs on \texttt{libero\_goal}, full injection (all 1024 dimensions) causes complete task failure (0\% success on 4/5 pairs, down from 100\% baseline). Goal-subspace injection (20 dimensions, 2\% of the activation space) preserves task success: 100\% on three pairs, 67\% on a fourth where the source and destination share the OPEN motor primitive. Action-subspace injection (15 dimensions encoding x/y/z action components) shows the same pattern. Replacing 2\% of the activation space from a different task does not disrupt the destination task's motor execution, evidence for separability of goal identity from motor execution within the same layer, and stops short of direct causal control over goal selection. Additive steering along goal directions in the VLM pathway is tested in App.~\ref{sec:silent-steering}.

\subsection{FFN Motor Primitive Neurons}
\label{sec:ffn-neurons}

Projecting PaliGemma FFN weight vectors onto a token-association vocabulary identifies neurons that encode specific motor primitives. At layer~17, 6,606 of 16,384 neurons (40.3\%) associate with motion verbs (pick, place, move, push, pull), 2,907 (17.7\%) with object tokens, 2,989 (18.2\%) with gripper state, 2,223 (13.6\%) with direction, and 1,151 (7.0\%) with spatial relations.

As we move through the layers we see the share of these neurons change. Motion verb neurons go from 4,231 at layer 0 (25.8\%) to 6,606 at layer 17 (40.3\%), a 56\% increase. Gripper neurons triple from 998 (6.1\%) to 2,989 (18.2\%). Object neurons decrease from 3,660 (22.3\%) to 2,907 (17.7\%), and spatial neurons decrease from 1,655 (10.1\%) to 1,151 (7.0\%). Going from objects/spatial to motion/gripper as we move through the layers shows how the model goes from what it sees to what it should do within one pathway.

OFT's 32-layer Llama-2 shows different characteristics. Concept mapping at finer granularity (80 concepts per layer across 11,008 neurons) shows spatial relation neurons dominating at every depth (\texttt{spatial\_in}: 5,483 $\to$ 7,245; \texttt{spatial\_on}: 4,724 $\to$ 6,727). These grow monotonically from layer 0 to layer 31, while neurons representing objects stay sparse (e.g., \texttt{obj\_can}: 814 $\to$ 1,175). Consistent with OFT's resilience to concept ablation (92\% of ablations had no effect), spatial information is spread across the layers of its 4096-dimensional representation. GR00T's SAE features show a gradient in feature use by layer type, with later DiT layers using more features than earlier ones. At DiT layer 0 only 3,654 out of 12,288 features are alive (70\% dead). By DiT layer 15 this number increases to 9,317 alive features. Eagle layer 0 has 6,730 alive features out of 16,384 and VL-SA layer 0 has 11,585 alive features.

\subsection{Layer Similarity via CKA}
\label{sec:cka}

Linear Centered Kernel Alignment (CKA) between \pizhalf{} expert layers 5 and 11 on LIBERO-Goal, computed on paired samples (the same rollout steps and token positions in both layers), is 0.79. A shuffled-row control gives 0.02. The expert therefore carries a shared residual representation that later layers refine, consistent with injected activations persisting through the remaining layers.

When looking at the probes on the 32 layers of the Llama-2 backbone used in OFT we see similar behavior. $R^2$ for predicting episode length increases from 0.845 at layer 0 to 0.941 at layer 24 then decreases to 0.915 at layer 31. Probes for classifying the task reach 97.7--100\% by layer 8. This, along with the previous results on the FFN layers, indicates that both architectures have layers that refine a shared representation.

\subsection{Where Spatial Binding Forms During Fine-Tuning}
\label{sec:base-vs-finetuned}

Comparing backbone activations between base OpenVLA (fine-tuned with discrete action tokens) and OFT (fine-tuned with continuous L1 regression) on the same LIBERO tasks localizes the representational changes induced by fine-tuning. We feed identical images through both models and measure per-layer cosine similarity of mean activations ($n{=}50$ images per suite, 4 suites).

We find that as we go deeper into the network the divergence increases, across all four suites. L1--L15 have cosine similarities of $>$0.999. L16--L28 range from 0.997 to 0.980 and the last layer L31 is 0.917--0.938. Layer 0 is slightly lower at 0.970--0.989 due to the difference in input embeddings.

The long-horizon \texttt{libero\_10} suite diverges most at L31 (0.917), and the three single-concept suites least (0.935--0.938).

Fine-tuning therefore reshapes the final third of the backbone and leaves early and middle layers intact. The coordinate-bound programs of \S\ref{sec:cross-task} sit in the layers that change most.

Four OFT models fine-tuned on different suites (6 pairs $\times$ 32 layers, $n{=}30$ images each) differ much less at L31 (0.973--0.992) and not at all in the middle (L16: 0.998--0.999). The action head (discrete against continuous) reshapes late layers more than the choice of fine-tuning suite.

\section{Goal Steering in the VLM Pathway}
\label{sec:silent-steering}

\S\ref{sec:pathway-specialization} places goal selection in the VLM pathway. Here we test that claim causally by asking whether an edit there, with no instruction present, makes the policy complete a chosen task in a scene that holds several possible goals.

\paragraph{Protocol.} We replace the instruction with a filler string of function words that have the same number of tokens, so all other tokens in the prompt keep their positions. We take the residual stream from the VLM pathway when given the true instruction and the filler instruction, with 3 rollouts for each. Only the instruction tokens are edited (and all tokens after the instruction for causal backbones). We try 3 types of edits. (i) A \emph{patch} replaces the edited tokens with their mean under the true instruction. (ii) A \emph{per-position gap} adds the per-position mean difference between the true instruction and filler. (iii) A \emph{single direction} adds a single unit vector, the per-position difference averaged over positions, scaled to a fraction $\rho$ of each token's norm. We use a random direction with the same norm and a gap from another task as controls. Filler and true instruction with no edit act as our floor and ceiling.

\paragraph{Result.} By editing one layer in the VLM pathway we are able to restore goal selection with no instruction on \pizhalf{}, SmolVLA, X-VLA and GR00T. When we edit with a random direction or a direction from another task of the same size we do not see this effect (Tab.~\ref{tab:silent-steering}). OpenVLA-OFT does not show this behavior. On \pizhalf{} we can complete 113/150 episodes with one direction at layer 2 of PaliGemma, compared to 6/150 with a random direction, and 9 out of 10 tasks show this behavior. We found layers that work (2 and 5) after failing to find any at layers 9 and 13 in our initial pilot, so the instruction is read early in the VLM stack.

The form of the signal depends on the architecture. \pizhalf{} carries the goal as one direction shared across instruction positions. SmolVLA carries it per position, so one averaged direction fails there at every magnitude. X-VLA's bidirectional Florence encoder spreads the instruction into the image tokens, so the edit has to cover the whole Florence block. GR00T responds to the per-position gap on its Eagle backbone at its ceiling. OpenVLA-OFT runs image and instruction through one Llama stack with bidirectional attention, and the per-position gap over the whole sequence at layer 4 or 12 completes 2/24 episodes on the eight tasks without a filler floor, against 24/24 for the true instruction.

\begin{table}[!htbp]
\centering
\caption{Goal steering with the instruction replaced by a length-matched filler, as successes over episodes for the chosen task. \pizhalf{}, GR00T (Eagle L6) and OFT (Llama L12, public checkpoint, official harness) run 10 \texttt{libero\_goal} tasks. X-VLA runs 9, excluding one task that succeeds under the filler. SmolVLA runs 5 \texttt{libero\_object} tasks.}
\label{tab:silent-steering}
\footnotesize
\begin{tabular}{@{}lccccc@{}}
\toprule
 & \pizhalf{} & SmolVLA & X-VLA & GR00T & OFT \\
\midrule
Filler, no edit            & 12/150 & 0/50  & 2/90  & 3/100 & 6/30 \\
True instruction           & 139/150 & 46/50 & 86/90 & 91/100 & 30/30 \\
Patch (instruction states) & 137/150 & 44/50 & --    & --     & -- \\
Per-position gap           & --      & 26/50 & 50/90 & 93/100 & 8/30 \\
Single direction           & 113/150 & 0/150 & --    & --     & -- \\
Random, matched norm       & 6/150   & 0/150 & 3/90  & 10/100 & 6/30 \\
Other task's edit          & 2/150   & 1/150 & 0/90  & 0/100  & 0/30 \\
\bottomrule
\end{tabular}
\end{table}

\paragraph{Pooling and SAE features.} We use the same setup to see if SAEs preserve the signal we can steer with (Tab.~\ref{tab:silent-steering-sae}). For X-VLA and GR00T we have per-token and mean-pooled SAEs on the layers we edit, and we add the per-position gap after applying each SAE. The per-token SAE reconstructs each position and the mean-pooled SAE reconstructs the mean of the positions, since that is what it was trained to do. On X-VLA we find that the per-token SAE can steer just as well as the gap itself (54/90 vs 50/90 at layer 0, Fisher $p{=}0.65$). The mean-pooled SAE falls to the random floor (2/90, $p{=}7.5{\times}10^{-19}$ compared to per-token). If we mean pool the raw gap without any SAE we already lose most of the ability to steer. We see similar results on GR00T. \S\ref{sec:sae-analysis} shows that X-VLA and the action head on GR00T can be mean pooled when reconstructing rollouts, but the instruction signal in their language pathway requires the per-token SAE. So pooling preference also depends on which signal the SAE needs to keep. On \pizhalf{} we can create a direction using the top 32 features from the per-token PaliGemma SAE and get 64/150 at layer 5 compared to 74/150 with the dense direction. SmolVLA's SAEs are trained on the output of MLPs, which do not contain the instruction in earlier layers (patching these alone gets us 0/50), so we cannot perform this test.

\begin{table}[!htbp]
\centering
\caption{Steering through SAE reconstructions. The per-position instruction gap is added after reconstruction by each SAE, with $n{=}90$ (X-VLA) and $n{=}100$ (GR00T) per cell.}
\label{tab:silent-steering-sae}
\footnotesize
\begin{tabular}{@{}lcccc@{}}
\toprule
 & X-VLA L0 & X-VLA L2 & GR00T L2 & GR00T L6 \\
\midrule
Raw gap                              & 50 & 48 & 94 & 93 \\
Per-token SAE                        & 54 & 46 & 66 & 88 \\
Raw gap, pooled over positions       & 14 & 10 & 9  & 24 \\
Mean-pool SAE, per position          & 13 & 6  & 8  & 17 \\
Mean-pool SAE on the pooled gap      & 2  & 3  & 4  & 10 \\
Random, matched norm                 & 3  & 0  & 5  & 10 \\
\bottomrule
\end{tabular}
\end{table}

\paragraph{Scope.} The edit is computed from rollouts under the true instruction in the same scene, so it restores a goal the policy already executes. It supplies a goal where no instruction is present and does not test overriding a competing instruction. Success rates are in simulation, with 3 (OFT) to 15 (\pizhalf{}) episodes per task for each edit and magnitude. Rows that pool magnitudes (SmolVLA single direction and controls) sum over them.

\section{Additional Experimental Results}
\label{sec:additional-results}

\paragraph{Tie-inclusive override rates.}\label{sec:tie-inclusive}
Table~\ref{tab:tie-inclusive} reports override without the tie-band exclusion of Eq.~\ref{eq:override}, i.e., $\Pr[\Delta_t > 0]$ over all injection steps, for every model with saved trajectories. The \pizhalf{} and OFT rows use the action-vector-cosine metric of the main text, and the X-VLA and GR00T rows use the end-effector-velocity classifier. X-VLA under action-vector cosine including ties is 39.5\%. Cross-architecture comparisons should be read within one metric.

\begin{table}[!htbp]
\centering
\caption{Tie-inclusive override ($\Pr[\Delta_t{>}0]$). $n$ counts episodes with saved trajectories, so it can be smaller than in Tab.~\ref{tab:cross-task}. $^{\ast}$SmolVLA's per-episode classifier has no tie band, so its tie-inclusive and conditional rates coincide.}
\label{tab:tie-inclusive}
\footnotesize
\begin{tabular}{@{}lrccc@{}}
\toprule
\textbf{Model} & $n$ & \textbf{Tie-incl.\ override [95\% CIs]} & \textbf{Tie fraction} & \textbf{Metric} \\
\midrule
\pizhalf{} & 1{,}968 & 99.6\% [99.2, 99.8] & \phantom{0}0.3\% & action cosine \\
OFT        & 1{,}079 & 77.9\% [75.4, 80.3] & 20.9\% & action cosine \\
X-VLA      & 2{,}907 & 52.8\% [51.0, 54.6] & 32.7\% & EEF-velocity cosine \\
GR00T      & \phantom{0,}180 & \phantom{0}6.7\% [3.9, 11.3] & \phantom{0}5.6\% & EEF-velocity cosine \\
SmolVLA    & \phantom{0,}720 & 52.1\% [48.4, 55.7] & \phantom{0}0.0\%$^{\ast}$ & per-episode override \\
\bottomrule
\end{tabular}
\end{table}

\paragraph{\pizhalf{} null-prompt recovery and batch agreement.} The \pizhalf{} null and injection rows of Tab.~\ref{tab:null-injection} come from one matched batch (10 tasks $\times$ 5 seeds, $n{=}50$/suite, standard chunked action execution). Against that batch's own baselines (90.0\%, 92.0\% and 88.0\% on goal, object and spatial), injection recovers 88\%, 69\% and 62\% of the baseline-to-null gap (Fisher $p{=}1.1{\times}10^{-14}$, $2.6{\times}10^{-6}$ and $0.13$, injection against null). The $n{=}500$ sweep in Tab.~\ref{tab:counterfactual} (31.6\% object, 5.8\% goal, 55.0\% spatial) agrees with it on all three suites (object and goal inside the $n{=}50$ Wilson intervals; spatial 31/50 vs.\ 275/500, Fisher $p{=}0.37$).

\begin{figure}[t]
    \centering
    \includegraphics[width=\linewidth]{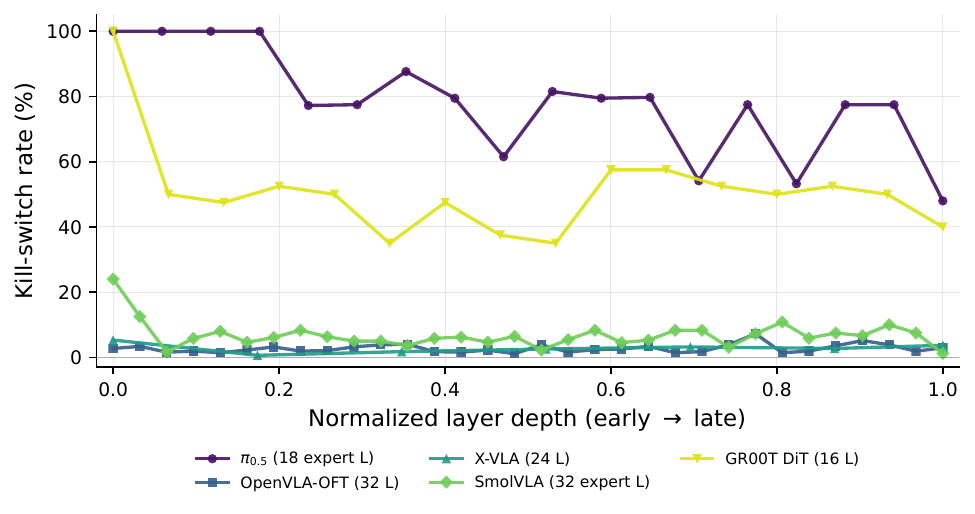}
    \caption{Per-layer kill-switch profile: the fraction of (concept, task) ablation pairs with $\Delta$SR $\geq 0.5$, against normalized layer depth. Pair counts are in Fig.~\ref{fig:concept-ablation-cross-model}.}
    \label{fig:cross-arch-killswitch}
\end{figure}

\paragraph{Steering sensitivity.} Models vary in their sensitivity to feature steering. \pizhalf{} expert pathways fail when suppressing features ($\Delta$SR $= -84$pp at $-3\times$) and GR00T DiT fails when amplifying features ($-68$pp at $9\times$). While OFT ($-6$pp at $-3\times$) and SmolVLA ($-3$pp at $5\times$) can handle these amounts of steering. We believe this is due to the sub-millimeter precision needed in manipulation tasks~\citep{chi2023diffusion}, errors accumulating over the 50 tokens of an action chunk~\citep{bengio2015scheduled}, and no closed-loop correction in behavior-cloned policies~\citep{dagger}. Steering during different task phases on \pizhalf{} shows that the model's sensitivity is localized to the transport phase ($p=0.013$, two-proportion $z$-test).

\subsection{Per-Token vs Mean-Pooled SAE Reconstruction}
\label{sec:pertoken-vs-meanpooled}

\begin{figure}[!htbp]
\centering
\includegraphics[width=0.92\linewidth]{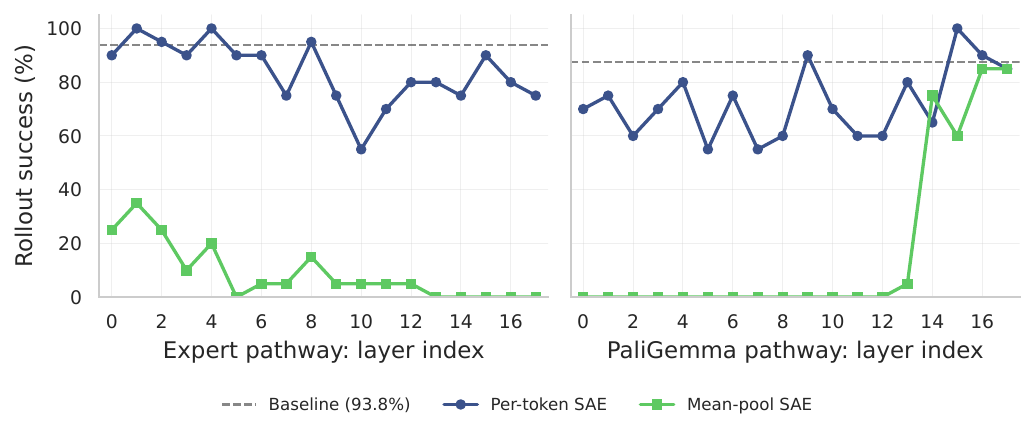}
\caption{Per-layer rollout success through per-token and mean-pool SAEs on \pizhalf{} \texttt{libero\_object} ($n{=}20$/cell). Dashed lines: per-pathway baselines (expert 93.8\%, PaliGemma 87.5\%).}
\label{fig:pertoken-vs-meanpool-pi05}
\end{figure}

Fig.~\ref{fig:pertoken-vs-meanpool-pi05} reports the 18-layer $\times$ 2-pathway $\times$ 2-pooling-mode rollout validation on \pizhalf{}/\texttt{libero\_object} ($n{=}20$ episodes per (layer, pooling) cell; 1,600 episodes in total). The two SAE families are trained on success-only baseline-rollout activations under identical hyperparameters (i.e., $k{=}64$, 8$\times$ expansion, 30 epochs), differing only in whether the latent code is computed per token or after mean-pooling and broadcast back to every position.

\begin{figure}[!htbp]
    \centering
    \includegraphics[width=0.72\columnwidth]{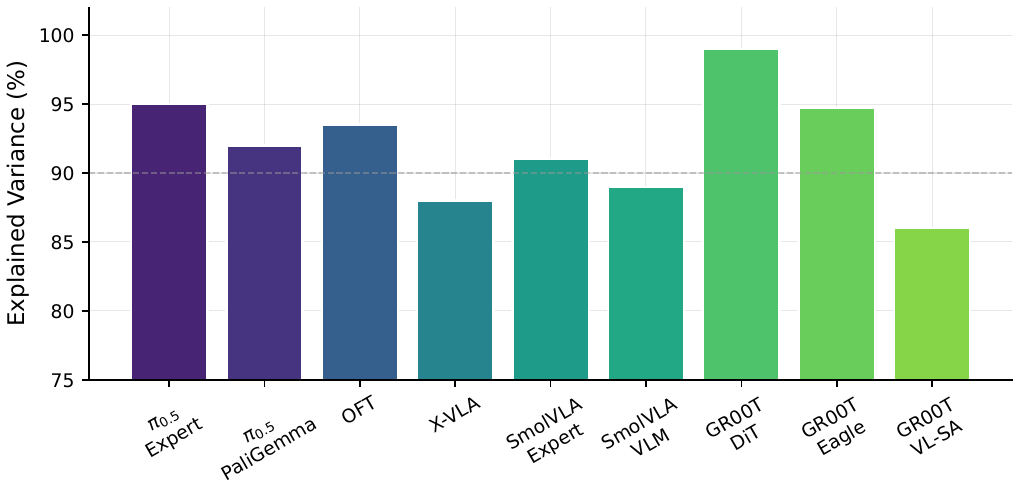}
    \caption{SAE explained variance by architecture and pathway (per-token SAEs, best layer). Dashed line: $R^2{=}0.90$.}
    \label{fig:sae-ev}
\end{figure}

Fig.~\ref{fig:sae-ev} shows cross-model SAE reconstruction. Mean-pool destroys the \pizhalf{} expert at all layers, dropping from a baseline of 93.8\% to 8.9\% averaging over all 18 layers. Per-token maintains success at all expert layers (range 55--100\%, with an average of 83.6\% over 18 layers). For the PaliGemma pathway we see the opposite happen at the end of the stack. Mean-pooled SAEs are able to maintain 60--85\% success at layers 14--17 while failing at earlier layers (0\% success at layers 0--12 and 5\% at layer 13). This is consistent with PaliGemma's role as a goal/scene encoder whose late-layer activations do not vary by position and so tolerate pooling. We also compare each task in LIBERO-10 for the \pizhalf{} expert ($n{=}250$ for each task) and find similar but more pronounced results: the per-token SAE achieves 1{,}748/2{,}500 successes (69.9\%) and mean-pool 9/2{,}500 (0.4\%).

\subsection{Temporal-Contrastive SAE Pooling (T-SAE Ablation)}
\label{sec:temporal-sae-arm}

Does adding a temporal-aware SAE that operates per token but pushes neighboring codes to be closer together close the per-token vs. mean-pool gap? We take the Temporal SAE (T-SAE) recipe from \citet{bhalla2025tsae} and apply it to VLA action tokens. Using this as a third pooling arm, we compare to the per-token and mean-pool results from Fig.~\ref{fig:pertoken-vs-meanpool-pi05} with the same activations. TopK ($k=64$), expansion ($8\times$), optimizer and number of epochs are all the same. See Training details below for the loss and other training settings.

\paragraph{Pooling preference is architecture-tied.} T-SAE checkpoints are trained on five architectures (\pizhalf{} expert 18 layers, X-VLA 24, SmolVLA expert 32, GR00T-DiT 16, OFT 32) and validated by rollout on all five (Tab.~\ref{tab:cross-arch-3arm-rollout}). Two regimes appear. On \pizhalf{} ($n{=}1{,}440$ per arm, Tab.~\ref{tab:temporal-sae-3arm}) and OFT (Tab.~\ref{tab:oft-3arm-per-suite}), mean-pooling collapses rollout success, so both models carry position-distinct phase information in the action-token residual stream. X-VLA ($n{=}4{,}800$ per arm), SmolVLA, and GR00T-DiT carry close to position-invariant action-token activations and tolerate mean-pooling, with SmolVLA indifferent within 0.8\,pp. OFT's similarity to \pizhalf{}, despite a different VLM and action head, places the divider in the residual-stream encoding. The temporal-contrastive arm degrades both regimes (mechanism below).

\begin{table}[!htbp]
\centering
\caption{\pizhalf{} expert 3-arm SAE pooling ($n{=}1{,}440$/arm, 18 layers $\times$ 4 suites $\times$ 20 ep). Deltas vs.\ Tab.~\ref{tab:null-injection} baseline.}
\label{tab:temporal-sae-3arm}
\footnotesize
\begin{tabular}{@{}lcccc@{}}
\toprule
\textbf{Suite} & \textbf{Baseline} & \textbf{Per-token} & \textbf{Temporal} & \textbf{Mean-pool} \\
\midrule
libero\_goal     & 93.2\% & 83.1\% (\textminus10.1) & 76.1\% (\textminus17.1) & \phantom{0}8.1\% (\textminus85.1) \\
libero\_spatial  & 92.8\% & \textbf{90.0\%} (\textminus2.8) & \textbf{88.3\%} (\textminus4.5) & \phantom{0}1.9\% (\textminus90.9) \\
libero\_object   & 87.6\% & 81.9\% (\textminus5.7)  & 74.7\% (\textminus12.9) & \phantom{0}2.2\% (\textminus85.4) \\
libero\_10       & 86.0\% & 70.3\% (\textminus15.7) & 46.7\% (\textminus39.3) & \phantom{0}1.1\% (\textminus84.9) \\
\midrule
\textbf{Mean (4 suites)} & \textbf{89.90\%} & \textbf{81.32\%} & \textbf{71.46\%} & \textbf{3.33\%} \\
\bottomrule
\end{tabular}
\end{table}

\paragraph{Mechanism.} We run a diagnostic on layer 8 (1.5M pairs of adjacent timesteps with 5K held out) to see how close the codes $z = \mathrm{encode}(\mathbf{h})$ of adjacent timesteps are to one another. Per-token and mean-pool have $\cos(z_t, z_{t+1})$ of 0.857 and 0.852 respectively, meaning that pooling does not affect this property and its damage comes at inference. The temporal-contrastive variant is able to increase this to 0.997, which is the optimal value for its InfoNCE loss. Because adjacent timesteps already share most of the same active features (TopK Jaccard of 0.61--0.71 compared to 0.12--0.18 for random pairs), the contrastive loss only needs to make the coefficients similar. This causes the adjacent timestep codes to become nearly identical and lose phase information that the policy uses later.

\paragraph{Cross-architecture confirmation.} The contrastive arm converges to $\cos_\text{adj} \approx 0.997$ on \pizhalf{}, X-VLA, SmolVLA, and GR00T-DiT (SmolVLA 0.940, slightly under-converged) while keeping per-token-grade R\textsuperscript{2} ($\geq 0.97$). On OFT, training is data-limited ($\sim$26K pairs/layer vs.\ $\geq 100$K elsewhere) and $\cos_\text{adj}$ rises only to 0.717. Per-architecture per-token reconstruction means and InfoNCE values are folded into Tab.~\ref{tab:cross-arch-3arm-rollout}.

\begin{table}[!htbp]
\centering
\caption{Rollout fidelity of the three SAE arms per architecture, with T-SAE training metrics under matched hyperparameters ($\alpha{=}1.0$, $k{=}64$, $8\times$). \textbf{Bold}: best arm per row.}
\label{tab:cross-arch-3arm-rollout}
\footnotesize
\resizebox{\columnwidth}{!}{%
\begin{tabular}{@{}llccccccc@{}}
\toprule
\textbf{Model (pathway)} & \textbf{L} & \textbf{Baseline} & \textbf{Per-token} & \textbf{Temporal} & \textbf{Mean-pool} & \textbf{Recon (PT)} & \textbf{Recon (T)} & \textbf{InfoNCE} \\
\midrule
\pizhalf{} (expert)         & 18 & 89.9\% & \textbf{81.3\%} & 71.5\% & \phantom{0}3.3\% & 0.018 & 0.026 & 5.26 \\
X-VLA (transformer)         & 24 & 95.2\% & 50.6\%          & 23.5\% & \textbf{80.4\%} & 0.008 & 0.015 & 5.26 \\
SmolVLA (expert)            & 32 & 67.3\% & \textbf{67.2\%} & 35.1\% & 66.4\%          & 0.002 & 0.108 & 5.35 \\
OFT (Llama-2)               & 32 & 95.4\% & \textbf{80.1\%} & 46.2\% & \phantom{0}7.3\% & 0.072 & 0.342 & 5.55 \\
GR00T (DiT)                 & 16 & 83.9\% & 80.0\%          & 76.2\% & \textbf{82.6\%} & 0.010 & 0.008 & 5.28 \\
\bottomrule
\end{tabular}}
\end{table}

\paragraph{Limitations.} We did not sweep $\alpha$, and use the default $\alpha=1.0$ published by \citet{bhalla2025tsae}. Thus the conclusion is that, at the published hyperparameters, T-SAE degrades rollout fidelity on all the VLA architectures we tested. It remains open whether a tuned $\alpha < 1$ may encourage some temporal smoothness without causing the latent codes to collapse (see Mechanism).

\paragraph{Training details.}\label{sec:tsae-details}
We follow the T-SAE recipe of \citet{bhalla2025tsae}, but use TopK selection with $k{=}64$ in place of its BatchTopK $k{=}20$, and omit its Matryoshka decoder. Hence the three arms differ only in pooling and in the inclusion of the contrastive term. In addition to the per-token reconstruction MSE, the total loss includes a symmetric InfoNCE term between adjacent latent codes:
\begin{equation}
\mathcal{L} \;=\; \mathcal{L}_{\text{recon}} \;+\; \alpha \, \mathcal{L}_{\text{contr}},\qquad
\mathcal{L}_{\text{contr}} \;=\; -\tfrac{1}{N}\sum_i \log \frac{\exp s(z_t^i, z_{t-1}^i)/\tau}{\sum_j \exp s(z_t^i, z_{t-1}^j)/\tau} \;+\; \text{(symmetric)},
\label{eq:tsae-loss}
\end{equation}
where $s(\cdot,\cdot)$ is cosine similarity, $\alpha=1.0$ and temperature $\tau=1.0$. VLA activations have shape $T \times S \times D$, corresponding to timesteps, tokens per step and dimension respectively. Latents at consecutive timesteps were considered adjacent, that is, \texttt{cross\_time} adjacency $(t, s) \leftrightarrow (t+1, s)$, the most direct analogue of adjacent tokens in T-SAE's InfoNCE.

In all three arms, we used the same activations restricted to successful rollouts only, $8\times$ expansion, Adam ($\text{lr}=10^{-3}$, batch~512, 100~epochs), and normalized each position separately. As in the original recipe, in-batch negatives may be other timesteps within the same episode.

\subsection{Dynamic Time Warping Analysis of Cross-Task Trajectories}
\label{sec:dtw-cross-task}

The \emph{cosine-dominance} override metric of \S\ref{sec:cross-task} (Eq.~\ref{eq:override}) compares the direction of the action vector at each step, which is relatively local, so we also wished to compare the overall shape of the trajectory. To this end we computed Dynamic Time Warping (DTW) on the end-effector position trajectories, divided by the length of the warping path. Given a cross-task pair $(A, B)$, with prompt~$B$ and source~$A$'s expert activations injected at layers~16--17, we obtain three DTW distances for every episode: $d_{\text{src}} = \mathrm{DTW}(\mathbf{x}^{B|A\text{-inj}}, \mathbf{x}^{A})$, $d_{\text{dst}} = \mathrm{DTW}(\mathbf{x}^{B|A\text{-inj}}, \mathbf{x}^{B})$, and the inherent $d_{A,B} = \mathrm{DTW}(\mathbf{x}^{A}, \mathbf{x}^{B})$ as a control. We are interested in the fraction of episodes where $d_{\text{src}} < d_{\text{dst}}$, the DTW source-dominance. This is directly analogous to the cosine-dominance override rate.

\input{dtw_table.tex}

DTW also indicates strong preference towards the source on \texttt{libero\_goal} and \texttt{libero\_spatial}, with source-dominance of 90.7\% ($n{=}270$) and 86.4\% ($n{=}44$) respectively. It is not surprising that this should be 8--12pp lower than the corresponding cosine numbers, since cosine measures the \emph{direction} of the action vector step by step while DTW measures the \emph{shape} of the whole trajectory: an injected episode may follow the source for a short period and then diverge, and so be cosine-dominant but mixed by DTW.

For the long-horizon \texttt{libero\_10} suite there is much less agreement, with only 45.8\% DTW source-dominance compared to 95.8\% cosine-dominance ($n{=}24$). The inherent $d_{A,B}{=}0.361$ is large, and injection often truncates the rollout, so that the truncated trajectory is artifactually ``closer'' in raw shape distance to the shorter destination baseline. Since cosine is normalized per step, it does not suffer from this effect. The \texttt{libero\_10} gap therefore reflects trajectory length, and there is no reason to doubt the cosine result.

\paragraph{Phase-aligned injection.} When injecting by step index we assume that the source and destination episodes are in phase. To test this we aligned source to destination steps using dynamic time warping on the end-effector trajectory (9 \texttt{libero\_goal} pairs $\times$ 6 episodes, $n{=}54$ per arm, expert L16+L17). We see 94.4\% override when using DTW compared to 100\% when using step index (Fisher $p{=}0.243$). If we shuffle the steps from the source episode we drop to 87.0\% ($p{=}0.013$), showing that we can detect if order matters. Warping with the same budget (88.9\%) and aligning on gripper phase (83.3\%) give similar results, with 0.0\% as our floor without injection. DTW returns the identity for 5 of the 9 pairs on this checkpoint. On the 3 checkpoints from App.~\ref{sec:datascale}, where alignment changes about 70\% of injected steps, we see 98.1--100\% override with step index, 66.7--68.5\% with DTW and 70.4--72.2\% with shuffling. Our floor is 1.9--5.6\%. On X-VLA we see 88.9\% override with DTW compared to 100\% with step index ($p{=}0.027$) and a floor of 0.0\% ($n{=}54$ per arm). Most of the override is maintained under any reordering of the source steps, and the amount that is lost is the same when aligning on phase and when shuffling.
\subsection{\pizhalf{} Concept-Feature Analysis}
\label{sec:layer-concept-specificity}

Tab.~\ref{tab:pi05-concept-features} reports which expert layers carry each concept and whether one feature separates the tasks that use it. Later expert layers (L14--L17) carry the strongest signal, and each top feature classifies its tasks at 100\% accuracy.

\begin{table}[!htbp]
\centering
\caption{\pizhalf{} concept features. (a) The three expert layers with the largest summed activation $\sum_t \mathrm{ReLU}(z_{f,t})$ per episode for each concept's top feature, in thousands. (b) Whether the top feature alone separates the LIBERO-10 tasks that use the concept.}
\label{tab:pi05-concept-features}
\small
\begin{tabular}{@{}llll@{}}
\toprule
\multicolumn{4}{@{}l}{\textit{(a) Layer specificity}} \\ \addlinespace[2pt]
\textbf{Concept} & \textbf{Best Layer} & \textbf{2nd Best} & \textbf{3rd Best} \\
\midrule
\multicolumn{4}{@{}l}{\textit{Action concepts}} \\
PUT & L17 (133k) & L16 (111k) & L15 (\phantom{0}86k) \\
OPEN & L15 (103k) & L16 (\phantom{0}79k) & L13 (\phantom{0}71k) \\
PUSH & L17 (412k) & L13 (275k) & L14 (270k) \\
INTERACT & L17 (274k) & L12 (261k) & L15 (253k) \\
\multicolumn{4}{@{}l}{\textit{Object concepts}} \\
BOWL & L16 (114k) & L15 (\phantom{0}80k) & L14 (\phantom{0}70k) \\
WINE\_BOTTLE & L16 (128k) & L14 (116k) & L17 (104k) \\
STOVE & L17 (182k) & L15 (164k) & L12 (141k) \\
\midrule
\addlinespace[4pt]
\multicolumn{4}{@{}l}{\textit{(b) Task classification on LIBERO-10}} \\ \addlinespace[2pt]
\textbf{Concept} & \textbf{Tasks Active} & \textbf{Tasks Inactive} & \textbf{Accuracy} \\
\midrule
PUT & 1,3,4,5,6,7,8 & 0,2,9 & 100\% \\
OPEN & 0,1 & 2--9 & 100\% \\
INTERACT & 9 & 0--8 & 100\% \\
\bottomrule
\end{tabular}
\end{table}

On \pizhalf{} libero-object, concept-aligned selection (Cohen's $d_f \times \mathrm{freq}_f$) preserves rollout success under ablation (93\% $\to$ 93\%), while action-correlated selection (linear-probe $x{,}y{,}z{,}\text{rpy}{,}\text{gripper}$) drops success by 14\,pp because the chosen feature tracks per-step motor output statistics and carries little concept identity.

\subsubsection*{Causal Necessity of Linear Probe Directions}
\label{sec:causal-features}
\begin{figure}[!htbp]
    \centering
    \includegraphics[width=0.85\columnwidth]{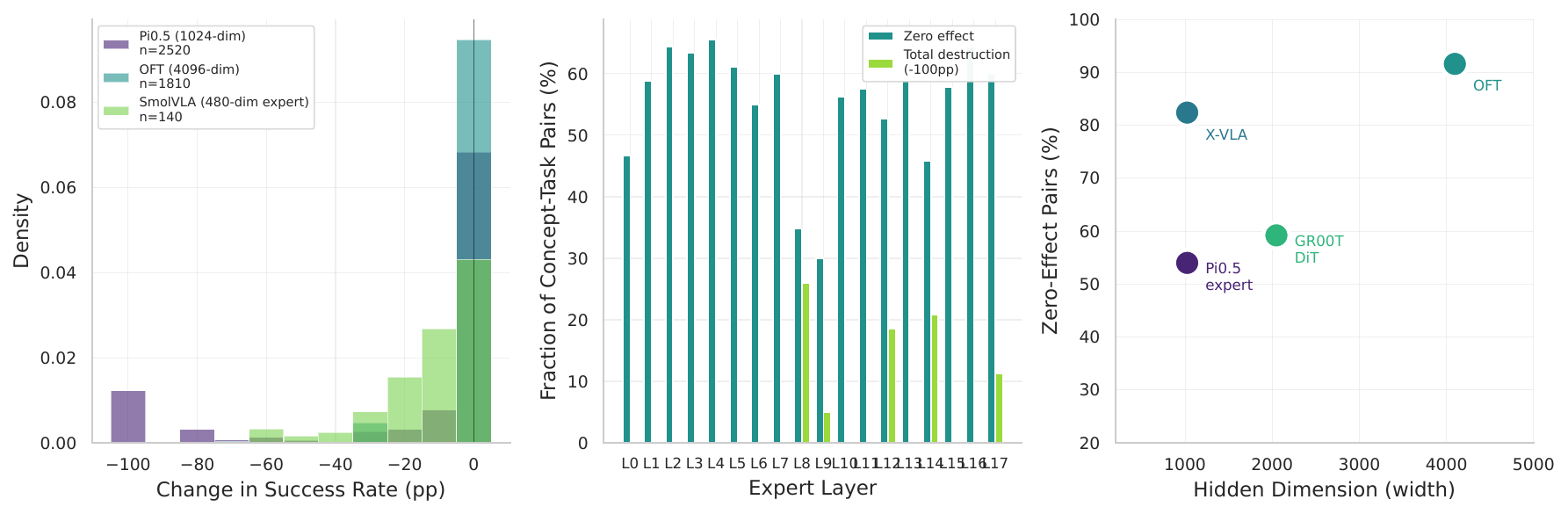}
    \caption{Single-feature ablation sensitivity. Left: change in success per (concept, task) pair for \pizhalf{} ($n{=}2{,}520$), OFT (1{,}810) and SmolVLA (140). Middle: per-layer fraction of zero-effect and total-destruction pairs on the \pizhalf{} expert. Right: zero-effect rate against hidden width for four models.}
    \label{fig:concept-ablation-cross-model}
\end{figure}

Linear probes trained to predict action dimensions achieve 97--98\% $R^2$. Projecting out probe directions eliminates linear decodability by construction ($R^2$ drops to $\approx 0$). The causal test is behavioral. Rollouts under projection fail (\S\ref{sec:linear-probes}), indicating the network itself relies on information in these directions. Cross-architecture single-feature ablation sensitivity is reported in Fig.~\ref{fig:concept-ablation-cross-model} ($n{=}15{,}096$ concept-task pairs across five models).

\subsubsection*{Vision Robustness}
Systematic image perturbation across 6,000+ episodes reveals task-dependent visual robustness. We apply horizontal and vertical flips, rotations ($90^\circ$, $180^\circ$, $270^\circ$), center crops (50\%, 75\%), and object-centric crops. Horizontal and vertical flips break all models tested (0\% success), so spatial orientation is rigidly encoded. Rotation and crop robustness varies by task complexity. Simple pick-and-place tolerates mild perturbations while multi-step tasks fail. Object-centric cropping (centering on the manipulation target) outperforms static crops (60\% vs.\ 0--20\%), pointing to reliance on manipulation-relevant regions, with full scene context contributing less.

\section{Ablation Studies}
\label{sec:ablation-studies}

\subsection{Temporal Ablation: GR00T Phase-Conditioned Feature Importance}
\label{sec:temporal-ablations}

Table~\ref{tab:groot-temporal} shows GR00T N1.5 temporal ablation aggregated over the full 32-layer stack across three LIBERO suites (160 conditions). Early-window ablation ($-29.7$pp pooled) is nearly as severe as full-episode ablation ($-30.5$pp), while mid- and late-window ablation cause only $-13.8$ and $-13.6$pp. The asymmetry holds on goal and long, and object stays near ceiling in every window. Layer-type-resolved profiles show DiT layers carry the asymmetry (early-window $-50$pp, similar to full-episode) while Eagle LM layers are flatter ($\sim -15$pp early). The effect scales with task complexity. \texttt{libero\_long} (303-step mean) shows a $-44$pp early drop vs.\ $-27$pp on \texttt{libero\_goal} (108-step mean). The per-layer-type decomposition is not separately tabulated here. GR00T is most sensitive during trajectory initiation, after which it tolerates feature ablation.

\begin{table}[!htbp]
\centering
\caption{GR00T N1.5 temporal ablation by LIBERO suite (160 conditions across 32 layers).}
\label{tab:groot-temporal}
\footnotesize
\begin{tabular}{@{}lccccc@{}}
\toprule
\textbf{Suite} & \textbf{Baseline} & \textbf{Full} & \textbf{Early} & \textbf{Mid} & \textbf{Late} \\
\midrule
LIBERO-Goal & 100\% & 73.1\% & 73.2\% & 92.7\% & 92.4\% \\
LIBERO-Long & 100\% & 53.8\% & 55.6\% & 75.3\% & 75.9\% \\
LIBERO-Object & 100\% & 93.9\% & 94.0\% & 95.0\% & 95.3\% \\
\midrule
\textbf{Pooled (160 cond.)} & \textbf{100\%} & \textbf{69.5\%} & \textbf{70.3\%} & \textbf{86.2\%} & \textbf{86.4\%} \\
\bottomrule
\end{tabular}
\end{table}

\section{Qualitative Results and Additional Figures}
\label{sec:qualitative-results}

\begin{figure}[!htbp]
    \centering
    \begin{subfigure}[t]{0.49\linewidth}
        \includegraphics[width=\linewidth]{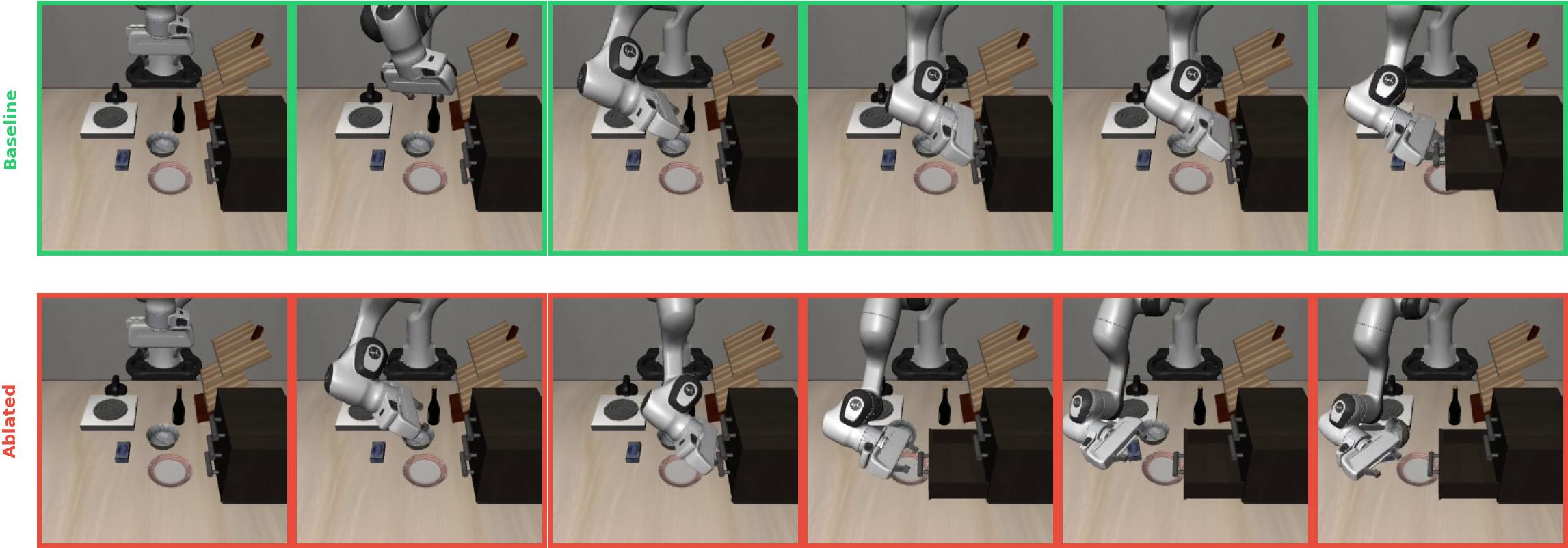}
        \caption{Open task (LIBERO).}
    \end{subfigure}
    \begin{subfigure}[t]{0.49\linewidth}
        \includegraphics[width=\linewidth]{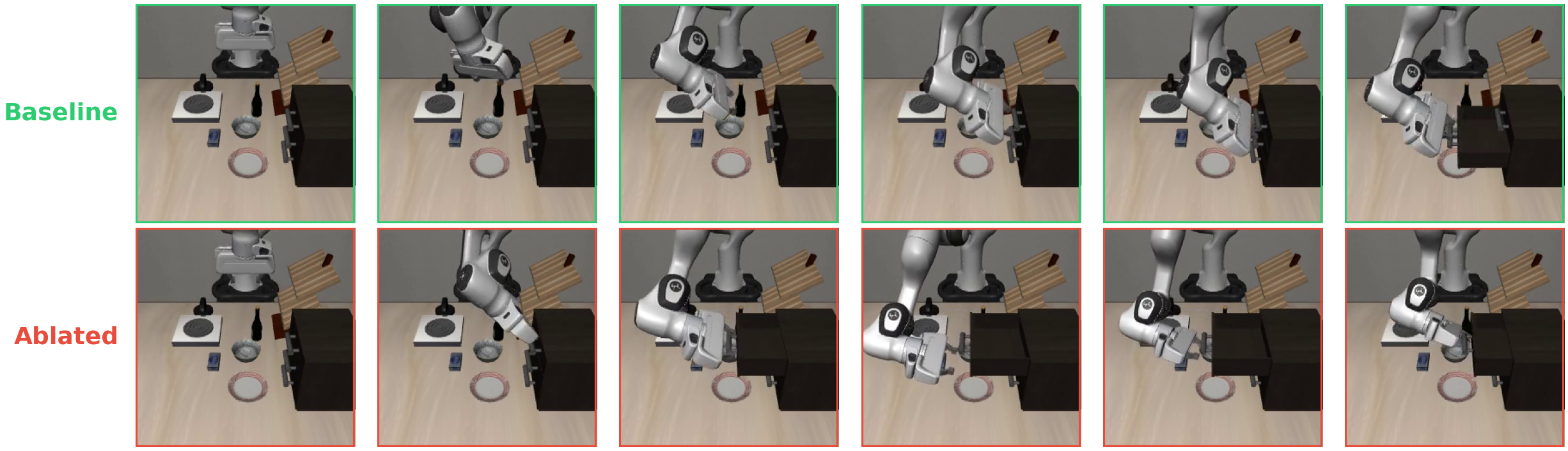}
        \caption{Push task (LIBERO).}
    \end{subfigure}
    \\[2pt]
    \begin{subfigure}[t]{0.49\linewidth}
        \includegraphics[width=\linewidth]{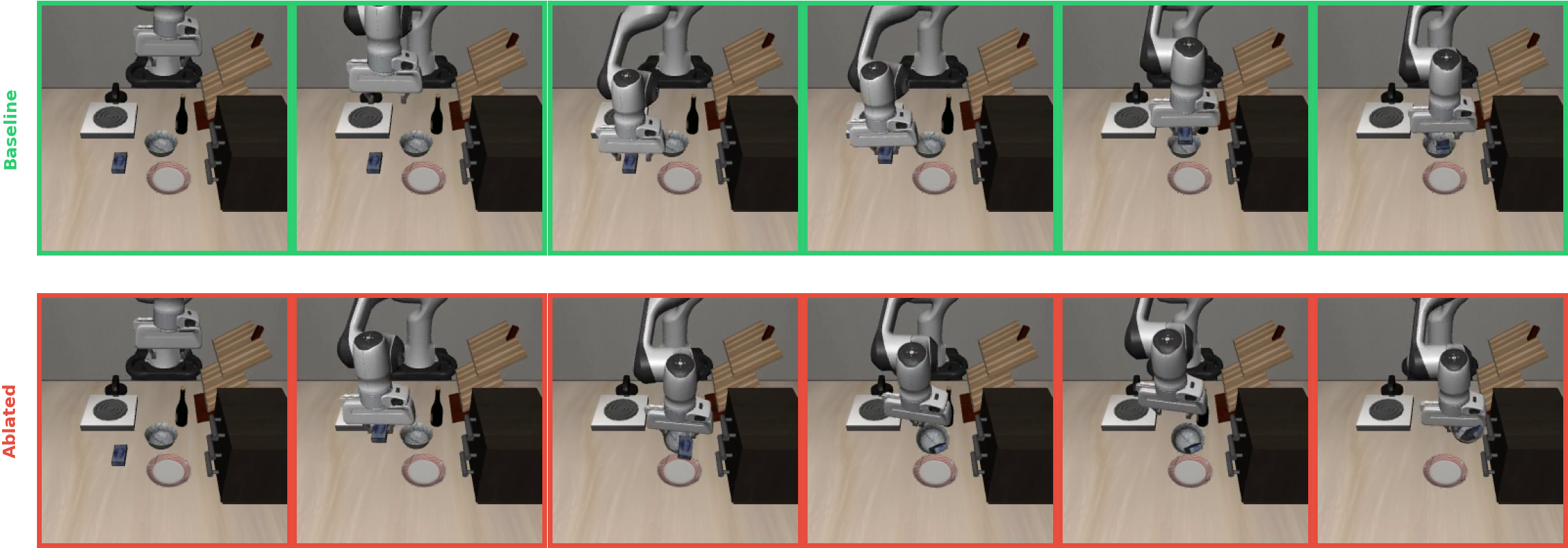}
        \caption{Put task (LIBERO).}
    \end{subfigure}
    \begin{subfigure}[t]{0.49\linewidth}
        \includegraphics[width=\linewidth]{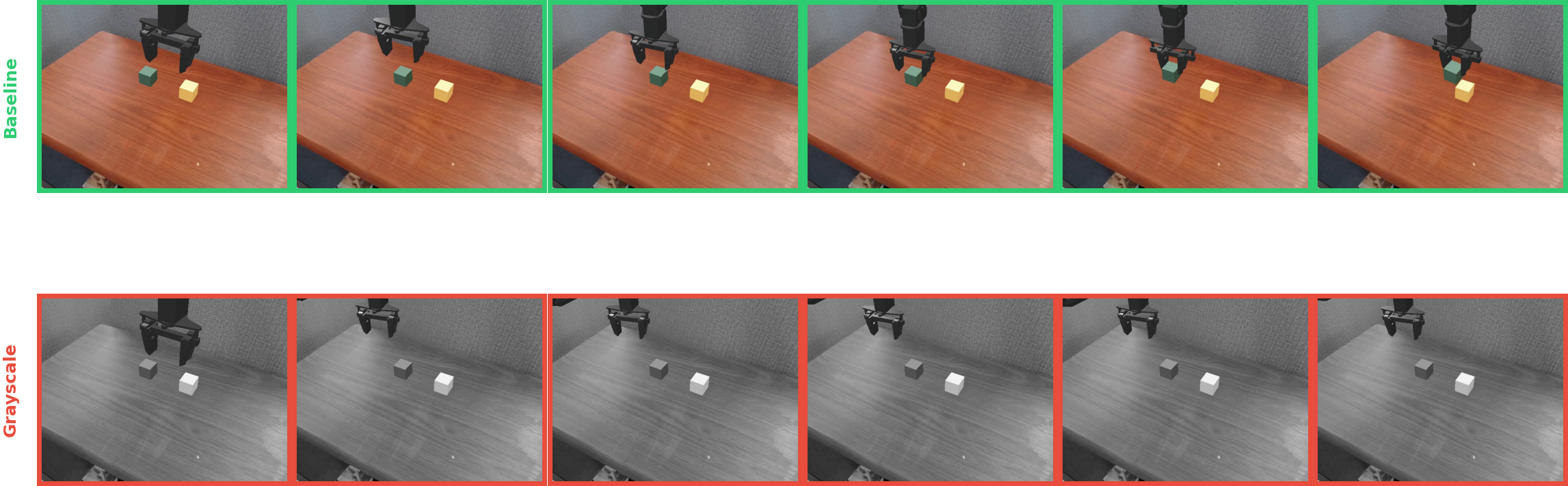}
        \caption{Grayscale perturbation (SimplerEnv WidowX).}
    \end{subfigure}
    \caption{Baseline (green) and ablated (red) task rollouts. (a)--(c) Feature ablation on LIBERO. (d) Grayscale input on SimplerEnv.}
    \label{fig:qualitative-rollouts}
\end{figure}

On SmolVLA and X-VLA, feature ablation and vision perturbation give binary outcomes: the robot completes the task or fails it, and failed episodes end in a partial reach. For example, SmolVLA 50\% center-crop removes button-press spatial context (400-step timeout vs.\ 67-step baseline), and X-VLA grayscale removes color discriminability for the WidowX stack-cube task (0\% success). Figure~\ref{fig:qualitative-rollouts} shows baseline and ablated rollouts. Rollout videos are on Action Atlas (\url{https://action-atlas.com}).

\begin{figure}[!htbp]
    \centering
    \includegraphics[width=0.88\columnwidth]{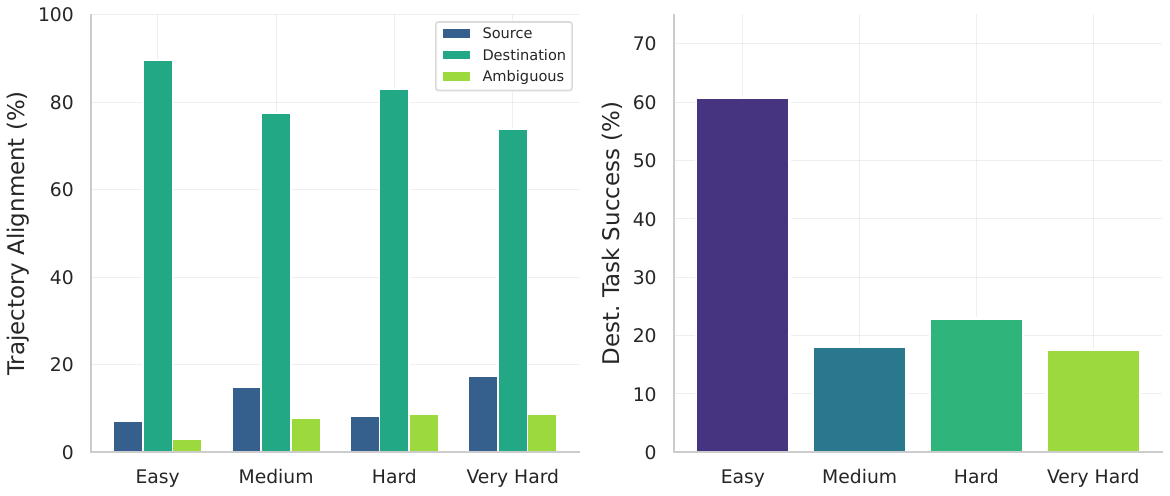}
    \caption{SmolVLA on MetaWorld by difficulty. Left: destination-behavior retention and source-task override under cross-task injection. Right: destination success under injection.}
    \label{fig:metaworld}
\end{figure}

\begin{figure}[!htbp]
    \centering
    \includegraphics[width=0.6\columnwidth]{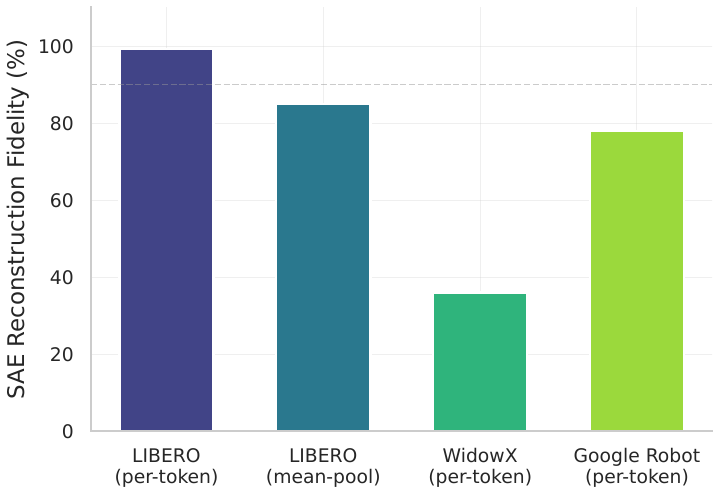}
    \caption{X-VLA SAE rollout fidelity across environments: success with the SAE reconstruction substituted at rollout time, relative to baseline.}
    \label{fig:simplerenv}
\end{figure}

\begin{figure}[!htbp]
    \centering
    \includegraphics[width=0.75\columnwidth]{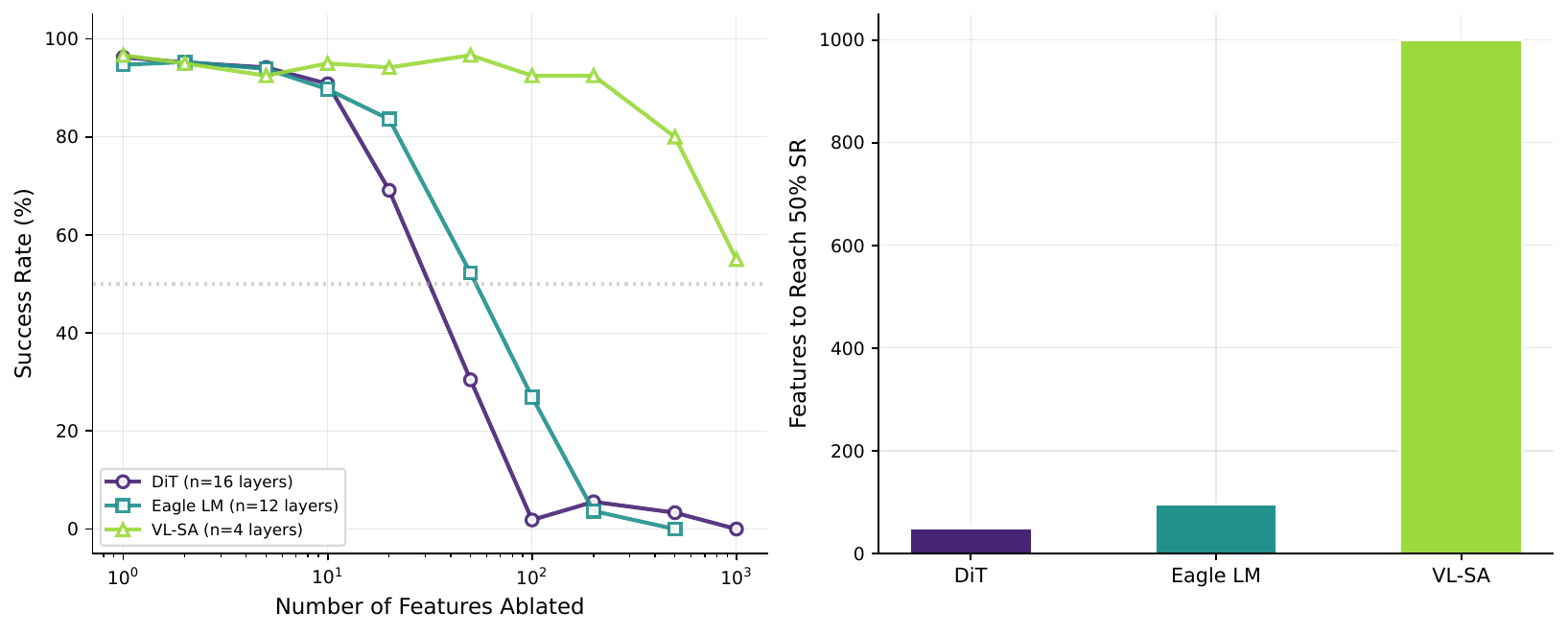}
    \caption{GR00T N1.5 success drop by layer type. DiT layers (16) drop 40--80\%, Eagle LM layers (12) drop moderately, and VL-SA layers (4) are the most resilient.}
    \label{fig:groot-layers}
\end{figure}

\begin{figure}[!htbp]
    \centering
    \includegraphics[width=0.55\columnwidth]{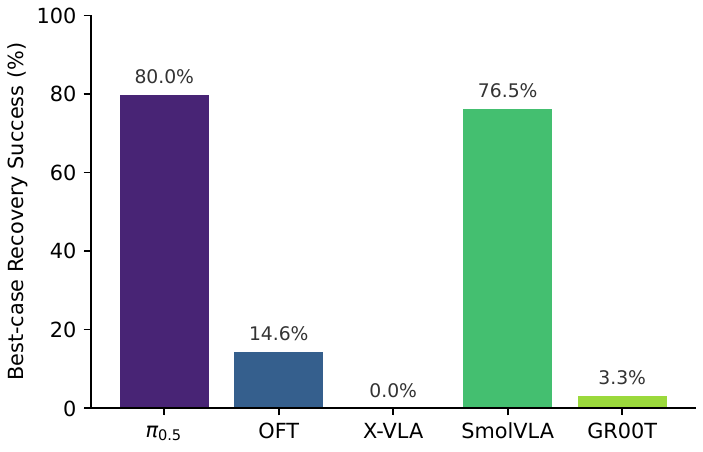}
    \caption{Best-case success under each architecture's visual-pathway intervention, from the rows of Tab.~\ref{tab:null-injection}: PaliGemma injection under a null prompt (\pizhalf{}), baseline injection under a null prompt (OFT), single-layer zeroing (X-VLA), single expert-layer zeroing on its best suite (SmolVLA), and null-source DiT injection (GR00T).}
    \label{fig:null-recovery}
\end{figure}

\section{Per-Architecture Deep Dives}
\label{sec:per-arch-deepdive}
\label{sec:per-suite-results}
\label{sec:per-arch-failures}

Each subsection gives the architecture's distinctive failure mode and its per-suite table. Baselines match Tab.~\ref{tab:null-injection} except where noted, and Fig.~\ref{fig:null-recovery} summarizes each architecture's best visual-pathway intervention. The ACT control is in App.~\ref{sec:act-control}.

\subsection{OpenVLA-OFT}
\label{sec:per-arch-oft}

OFT has the highest raw baseline (Tab.~\ref{tab:null-injection}) and achieves 23/25 on the \texttt{libero\_10} task-8 seed cluster where \pizhalf{} scores 0/25. Its main failure mode is structural. If we zero any single layer of the residual stream (0, 8, 16, 24 or 31) we see 0\% success on all suites, and if we inject the baseline activations from the visual pathway with a null prompt we see minimal recovery (Tab.~\ref{tab:oft-deep-dive}). OFT has the lowest null-injection recovery of the models tested. With temporal-window null injection on \texttt{libero\_10}, the nine tasks that succeed at baseline drop to 0\% in the early and mid windows. We are able to keep 2 tasks successful when nulling late. Episode-length probes are slightly stronger in the late layers for all suites (early/mid/late $R^2$: spatial 0.984/0.991/0.986, object 0.719/0.834/0.863, goal 0.859/0.896/0.933, \texttt{libero\_10} 0.733/0.779/0.827).

\begin{table}[!htbp]
\centering
\caption{OFT per-suite deep dive. Zero-layer: any of layers 0/8/16/24/31 nulled under empty prompt (baseline $\sim$90\%). Recovery: null + inject zero-baseline activations. Probe $R^2$: episode-length probe over 32 layers, $n{=}149$/suite.}
\label{tab:oft-deep-dive}
\footnotesize
\begin{tabular}{@{}lccccc@{}}
\toprule
& \multicolumn{2}{c}{\textbf{Null injection}} & \textbf{Same-scene} & \multicolumn{2}{c}{\textbf{Probe $R^2$ (32 L)}} \\
\cmidrule(lr){2-3}\cmidrule(lr){5-6}
\textbf{Suite} & \textbf{Zero-layer} & \textbf{Recovery} & \textbf{inj.\ SR} & \textbf{Min} & \textbf{Mean} \\
\midrule
libero\_goal    & $\sim$0\% & 33/234 (14\%) & 87.8\% (n=582)  & 0.845 & 0.896 \\
libero\_object  & $\sim$0\% & 43/294 (15\%) & 77.8\% (n=162)  & 0.608 & 0.813 \\
libero\_spatial & $\sim$0\% & 31/216 (14\%) & 74.0\% (n=192)  & 0.969 & 0.988 \\
libero\_10      & 0/10 tasks & --            & 71.8\% (n=582)  & 0.692 & 0.780 \\
\bottomrule
\end{tabular}
\end{table}

\paragraph{SAE rollout per suite.}
\label{sec:oft-groot-per-suite}

\begin{figure}[!htbp]
\centering
\includegraphics[width=0.85\columnwidth]{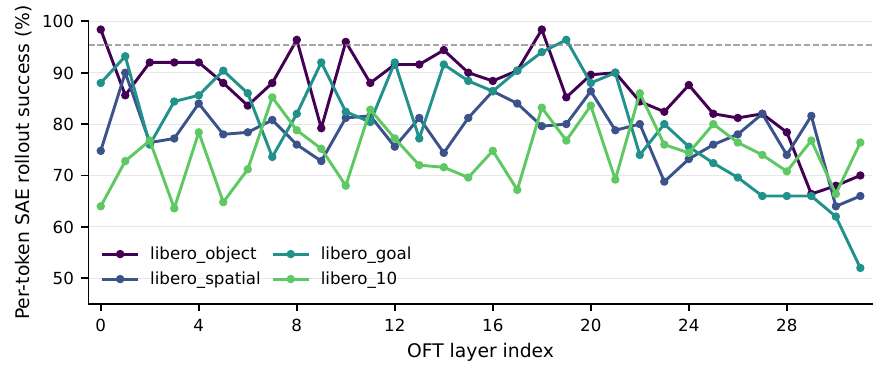}
\caption{OFT per-token SAE rollout success by decoder layer and LIBERO suite. Dashed line: rollout baseline (95.4\%).}
\label{fig:oft-pertoken-layerwise}
\end{figure}

The OFT means per suite are found in Tab.~\ref{tab:oft-3arm-per-suite}. The lower \texttt{libero\_10} per-token mean (74.5\%) is spread across layers (Fig.~\ref{fig:oft-pertoken-layerwise}), and all four suites decline after L20.

\begin{table}[!htbp]
\centering
\caption{OFT 3-arm SAE rollout per-suite (32 layers $\times$ 25 ep per cell). Per-token wins on every suite by 26--40pp, and mean-pool stays in single digits.}
\label{tab:oft-3arm-per-suite}
\footnotesize
\begin{tabular}{@{}lccc@{}}
\toprule
\textbf{Suite} & \textbf{Per-token} & \textbf{Temporal} & \textbf{Mean-pool} \\
\midrule
libero\_object  & \textbf{86.6\%} & 54.6\% & \phantom{0}9.2\% \\
libero\_spatial & \textbf{78.2\%} & 51.8\% & \phantom{0}6.9\% \\
libero\_goal    & \textbf{81.0\%} & 43.8\% & \phantom{0}8.0\% \\
libero\_10      & \textbf{74.5\%} & 34.5\% & \phantom{0}5.1\% \\
\midrule
\textbf{Mean (4 suites)} & \textbf{80.1\%} & 46.2\% & \phantom{0}7.3\% \\
\bottomrule
\end{tabular}
\end{table}

\label{sec:oft-pertoken-layerwise}

\label{sec:oft-per-suite}

\input{oft_per_suite_prose.tex}

\input{oft_per_suite_table.tex}

\subsection{Cross-Model Linear Probing}
\label{sec:cross-model-probing}

Task-relevant information is linearly decodable on every architecture (Tab.~\ref{tab:cross-model-probing}). Expert and action pathways encode state dynamics, and VLM pathways encode goal semantics.

\begin{table}[!htbp]
\centering
\caption{Cross-model linear probing. Oracle ratio: probe $R^2$ over oracle $R^2$, the fraction of ground-truth state linearly decodable from activations. The expert/VLM oracle split is measured on SmolVLA only, and \S\ref{sec:pathway-specialization} gives the \pizhalf{} and GR00T evidence.}
\label{tab:cross-model-probing}
\footnotesize
\begin{tabular}{@{}llcc@{}}
\toprule
\textbf{Model} & \textbf{Probe Target} & \textbf{Metric} & \textbf{Result} \\
\midrule
\pizhalf{} (expert) & State prediction & $R^2$ & 0.45 \\
\pizhalf{} (PaliGemma) & Prompt classification & Accuracy & 99.3\% \\
\midrule
OFT & Task identification & Accuracy & 97.7--100\% \\
OFT & Success prediction & AUC & 0.97 \\
OFT & Episode-length (mean) & $R^2$ & 0.87 \\
\midrule
SmolVLA (expert) & Oracle state ratio (h=10) & ratio & 0.58 \\
SmolVLA (VLM) & Oracle state ratio (h=10) & ratio & 0.13 \\
\midrule
GR00T (all layers) & Task identification & Accuracy & 100\% \\
GR00T (all layers) & Success prediction & Accuracy & 96.4\% \\
GR00T (DiT L14) & Success prediction & Accuracy & 97.7\% \\
\bottomrule
\end{tabular}
\end{table}

\subsection{\pizhalf{}}
\label{sec:per-arch-pi05}

\pizhalf{} achieves 89.9\% on the 1,000-rollout baseline (Tab.~\ref{tab:null-injection}). The main cause of failure is coordinate-cluster brittleness. For example, on \texttt{libero\_10} task 8 there is one range of 25 consecutive seeds that gets a 0/25 success rate, while the neighboring ranges get 23/25 and 20/25 on the same task with the same policy. This is a manifestation of the coordinate-bound program from \S\ref{sec:cross-task}. With the LIBERO-PRO position perturbations of \citet{zhou2025liberopro}, \pizhalf{} drops from 0.92--0.98 to 0.08--0.38 across all four suites.

\paragraph{Object identity does not predict reach target.} Changing the object name while keeping the scene fixed did not change the final end-effector position. One-way ANOVA of the final XYZ position by object name: \texttt{libero\_object} $F(10,209){=}1.27$, $p{=}0.247$, $\eta^2{=}0.057$ ($n{=}220$); \texttt{libero\_goal} $F(9,170){=}0.30$, $p{=}0.973$, $\eta^2{=}0.016$ ($n{=}180$); \texttt{libero\_spatial} $F(5,54){=}2.20$, $p{=}0.067$, $\eta^2{=}0.169$ ($n{=}60$). None of the three is significant at $p{<}0.05$, and the spatial test is underpowered ($n{=}60$). Combined with our results on null prompting, this shows that \pizhalf{} requires an object name to engage the task but does not use the object name to determine where to go.

\paragraph{Cross-task injection per suite.} When injecting into the model we see a decrease in destination success on every suite. \texttt{libero\_goal} goes from 91.5\% down to 0.0\% with a cross-prompt and no injection ($n{=}236$ for each condition; PaliGemma-all gets 0.4\% and expert L16 gets 1.3\%). \texttt{libero\_spatial} ($n{=}34$) drops from 100\% to 64.7\% with a cross-prompt and to 20.6\% with its own prompt and PaliGemma-all injection. \texttt{libero\_10} ($n{=}24$) drops from 62.5\% to 0\% for every injection condition.

\subsection{X-VLA}
\label{sec:per-arch-xvla}

X-VLA reaches 95.2\% on the baseline and matches the \pizhalf{} injection signature (\S\ref{sec:cross-task}), including the \texttt{libero\_10} task-8 seed cluster (0/25, with 24/25 on the neighboring range). Its failures are visual and all-or-nothing, with vision perturbations giving full success or 0\% (Fig.~\ref{fig:qualitative-rollouts}d). Its null-prompt success depends on the suite, and concept ablation barely moves it (Tab.~\ref{tab:xvla-per-suite}). Its SAEs prefer mean-pooling (80\% against 51\% per-token over 24 layers), so per-token SAEs lower X-VLA rollout success. The dictionaries are environment-specific (Fig.~\ref{fig:simplerenv}). At layer~0 the LIBERO SAE reconstructs LIBERO activations at normalized MSE 0.010 per-token and 0.006 mean-pool ($R^2{=}0.99$), and on WidowX activations it reaches MSE 0.66 ($R^2{=}0.34$), falling to $R^2{\leq}0$ by mid-depth.

\begin{table}[!htbp]
\centering
\caption{X-VLA per-suite breakdown. Zero any layer: any of 24 tested layers nulled. Null prompt: success under empty string. Zero eff./$\Delta$pp: concept ablation zero-effect rate and mean delta. $n{=}4{,}800$ counterfactual episodes and $n{=}2{,}480$ concept-ablation pairs.}
\label{tab:xvla-per-suite}
\footnotesize
\begin{tabular}{@{}lccccc@{}}
\toprule
\textbf{Suite} & \textbf{Baseline} & \textbf{Zero any layer} & \textbf{Null prompt} & \textbf{Zero eff.} & \textbf{$\Delta$pp} \\
\midrule
Goal & 98.8\% & 0\% & 10\% & 82.6\% & $-2.2$ \\
Object & 98.8\% & 0\% & 60\% & 98.2\% & $-1.3$ \\
Spatial & 95.6\% & 0\% & 48\% & 85.2\% & $-2.6$ \\
LIBERO-10 & 87.6\% & 0\% & 28\% & 74.7\% & $-2.4$ \\
\bottomrule
\end{tabular}
\end{table}

\subsection{SmolVLA}
\label{sec:per-arch-smolvla}

SmolVLA reaches 67.3\% on the baseline and fails in two ways. Tasks 0 and 4 of \texttt{libero\_10} sit far below the suite mean of 41.2\%. On the task-8 seed cluster it scores 0/25 and recovers only to 12\% and 60\% on the neighboring ranges. On MetaWorld, injection leaves SmolVLA's destination behavior mostly intact (89.7\% easy to 73.8\% very hard) and source override low (12.4\% aggregate, Fig.~\ref{fig:metaworld}). Expert concept ablation hurts \texttt{libero\_spatial} most, while VLM concept ablation hurts \texttt{libero\_goal} most and helps \texttt{libero\_object}, consistent with the VLM encoding goal semantics (Tab.~\ref{tab:smolvla-per-suite}).

\begin{table}[!htbp]
\centering
\caption{SmolVLA per-suite breakdown. Expert zero: success-rate range when zeroing individual expert layers. Zero eff./$\Delta$pp: expert-pathway concept ablation ($n{=}1{,}696$). VLM concept ablation ($n{=}210$): goal $-10.7$pp, object $+6.4$pp, spatial $-8.5$pp, LIBERO-10 $+5.9$pp.}
\label{tab:smolvla-per-suite}
\footnotesize
\begin{tabular}{@{}lcccc@{}}
\toprule
\textbf{Suite} & \textbf{Baseline} & \textbf{Expert zero} & \textbf{Zero eff. (exp)} & \textbf{$\Delta$pp (exp)} \\
\midrule
Goal & 74.8\% & 60--83\% & 14.6\% & $-4.7$ \\
Object & 87.6\% & 60--83\% & 0.0\% & $-3.9$ \\
Spatial & 65.6\% & 47--77\% & 32.7\% & $-17.7$ \\
LIBERO-10 & 41.2\% & 0--33\% & 20.3\% & $+0.8$ \\
\bottomrule
\end{tabular}
\end{table}

\subsection{GR00T N1.5}
\label{sec:per-arch-groot}

GR00T achieves 83.9\% on the baseline (Tab.~\ref{tab:null-injection}, spatial$^{\ast}$). Zeroing any one of the 16 DiT layers gives a 0\% success rate on all suites, while zeroing VL-SA layers keeps 40--80\% (Fig.~\ref{fig:groot-layers}), showing that the motor expert is doing most of the work. Similar to X-VLA, language sensitivity depends on the suite. \texttt{libero\_long} has the lowest concept-ablation zero-effect rate at 42.2\% and the highest destruction rate at 11.0\% (Tab.~\ref{tab:groot-per-suite}). The concept-ablation baselines in the table come from a separate batch ($n{=}10$ for each task).

\paragraph{Checkpoints.} For the goal results we use \texttt{Tacoin/GR00T-N1.5-3B-LIBERO-GOAL}, and for \texttt{libero\_10} we use \texttt{liorbenhorin-nv/groot-libero\_10-64\_40000}, the public checkpoints that had usable baselines (alternatives reach 10\% or less). $^{\ast}$ The spatial export was trained with a batch size of 128 for 20K steps, and the NVIDIA reference reports 97.65\% with a batch size of 640. Community fine-tunes range over 68--94\%, so we only report the spatial baseline (Tab.~\ref{tab:null-injection}, $^{\dagger}$). The qualitative findings hold on the 3 suites that have full coverage.

\begin{table}[!htbp]
\centering
\caption{GR00T N1.5 per-suite breakdown. Zero DiT: any of 16 DiT layers nulled. CF ``other'': counterfactual success under non-baseline prompts ($n{=}180$/suite). Zero eff./$\Delta$pp: concept ablation across 6{,}500 pairs.}
\label{tab:groot-per-suite}
\footnotesize
\begin{tabular}{@{}lccccc@{}}
\toprule
\textbf{Suite} & \textbf{Baseline} & \textbf{Zero DiT} & \textbf{CF ``other''} & \textbf{Zero eff.} & \textbf{$\Delta$pp} \\
\midrule
Goal & 97.0\% & 0\% & 18.9\% & 68.8\% & $-5.8$ \\
Object & 99.0\% & 0\% & 73.3\% & 69.6\% & $-9.9$ \\
Long & 75.0\% & 0\% & 61.7\% & 42.2\% & $-5.3$ \\
\bottomrule
\end{tabular}
\end{table}

\subsection{Operational Summary: Failure-Attribution Rules}
\label{sec:operational-summary}

These failures give three diagnostics. If the robot confidently goes to the wrong location, it is likely due to corruption in the expert pathway (\S\ref{sec:cross-task}). If it does not make a committed motion it is likely due to corruption in the VLM pathway (App.~\ref{sec:goal-encoding}). These apply to models that have separate pathways (\pizhalf{}, SmolVLA, GR00T). A failure on a contiguous range of seeds where the neighboring seeds pass points to brittleness around one layout of object coordinates, rather than a flaw in the whole model. It appears in models where absolute coordinates dominate the action distribution (\pizhalf{}, X-VLA, SmolVLA), and OFT is able to pass on these seeds.

\section{Dose-Response of Expert Injection}
\label{sec:dose-response}

The injection of Tab.~\ref{tab:cross-task} replaces the entire expert residual ($\alpha{=}1$), so it does not differentiate between a program that switches on at full magnitude and a bias that scales with the magnitude. A switch predicts destination behavior at every partial dose and a step at $\alpha{=}1$, and a bias predicts a graded rise as we scale up the magnitude. On \pizhalf{} \texttt{libero\_goal} we blend the output of expert layers 16/17 as $(1-\alpha)\,h_{\text{live}} + \alpha\,h_{\text{src}}$, where $h_{\text{src}}$ is the source activation from Eq.~\ref{eq:injection} recorded during one rollout without injection, and we keep the destination prompt fixed. We run 7 pairs of tasks with 5 episodes at each value of $\alpha \in \{0, 0.25, 0.5, 0.75, 1\}$. An episode is classified as source or destination when its action-trajectory cosine to the source rollout exceeds or trails its cosine to an uninjected destination rollout by more than $\tau{=}0.05$ (Eq.~\ref{eq:override}).

\begin{table}[!htbp]
\centering
\caption{Dose-response of expert injection on \pizhalf{} \texttt{libero\_goal}, 7 task pairs $\times$ 5 episodes per dose. Unanimous: pairs with all 5 episodes source-classified. Cosines are episode means.}
\label{tab:dose-response}
\footnotesize
\begin{tabular}{@{}lcccrr@{}}
\toprule
$\alpha$ & \textbf{Source-classified} & \textbf{Unanimous pairs} & \textbf{Dst.\ success} & $\overline{\cos}_{\text{src}}$ & $\overline{\cos}_{\text{dst}}$ \\
\midrule
0.00 & 0/35 = 0.0\%      & 0/7 & 34/35 = 97.1\% & 0.360 & 0.904 \\
0.25 & 4/35 = 11.4\%    & 0/7 & 19/35 = 54.3\% & 0.475 & 0.782 \\
0.50 & 34/35 = 97.1\%  & 6/7 & 3/35 = 8.6\%    & 0.705 & 0.312 \\
0.75 & 35/35 = 100\%    & 7/7 & 5/35 = 14.3\%   & 0.604 & 0.154 \\
1.00 & 35/35 = 100\%    & 7/7 & 4/35 = 11.4\%   & 0.551 & 0.147 \\
\bottomrule
\end{tabular}
\end{table}

Episodes that are classified as source go from 0/35 at $\alpha{=}0$ to 4/35 at 0.25 and 34/35 at 0.5 before plateauing. Mean cosines also cross in this range (Tab.~\ref{tab:dose-response}). Since the 0.5 dose alone rules out a step confined to $\alpha{=}1$, we read the effect as graded biasing of the expert's trajectory. At low doses we see a decrease in the ability to complete the instructed task (destination success 19/35 for $\alpha{=}0.25$) and then redirection at 0.5. Each dose has 35 episodes, which correspond to 7 clusters that share a source rollout. There are no pairs that are unanimously source below 0.5. At 0.5, 6/7 pairs are unanimous, and 7/7 are unanimous at 0.75 and 1.0. The 4 source episodes at 0.25 come from 2 pairs. This is limited to one checkpoint, task suite, and layer pair. Similar to the other simulation runs, we did not match sampling noise (App.~\ref{sec:physical}).

\section{SAE Sparsity Sweep}
\label{sec:sae-k-sweep}

The pooling comparison of \S\ref{sec:sae-analysis} uses $k{=}64$. We repeat both the per-token and mean-pool rollout on \pizhalf{} at $k \in \{32, 128\}$ at expert layers 5 and 11, on all four suites of LIBERO ($n{=}30$ per cell), using the replacement write-back of Fig.~\ref{fig:pertoken-vs-meanpool-pi05}. Each \texttt{libero\_10} arm has its own baseline run.

\begin{table}[!htbp]
\centering
\caption{Per-token and mean-pool SAE rollout success on \pizhalf{} across sparsity $k$, $n{=}30$ per cell, against the uninjected baseline of the same run. Per-token and mean-pool Wilson 95\% intervals are disjoint in every cell. Mean-pool uses the replacement write-back.}
\label{tab:sae-k-sweep}
\footnotesize
\begin{tabular}{@{}llcccc@{}}
\toprule
\textbf{Suite} & \textbf{Layer} & $k$ & \textbf{Baseline} & \textbf{Per-token} & \textbf{Mean-pool (replacement)} \\
\midrule
\texttt{libero\_goal}    & L5  & 32  & 28/30 = 93.3\% & 19/30 = 63.3\% & 0/30 = 0.0\% \\
\texttt{libero\_goal}    & L5  & 128 & 28/30 = 93.3\% & 29/30 = 96.7\% & 4/30 = 13.3\% \\
\texttt{libero\_goal}    & L11 & 32  & 28/30 = 93.3\% & 15/30 = 50.0\% & 2/30 = 6.7\% \\
\texttt{libero\_goal}    & L11 & 128 & 28/30 = 93.3\% & 28/30 = 93.3\% & 3/30 = 10.0\% \\
\texttt{libero\_object}  & L5  & 32  & 26/30 = 86.7\% & 19/30 = 63.3\% & 0/30 = 0.0\% \\
\texttt{libero\_object}  & L5  & 128 & 26/30 = 86.7\% & 24/30 = 80.0\% & 1/30 = 3.3\% \\
\texttt{libero\_object}  & L11 & 32  & 26/30 = 86.7\% & 13/30 = 43.3\% & 0/30 = 0.0\% \\
\texttt{libero\_object}  & L11 & 128 & 26/30 = 86.7\% & 26/30 = 86.7\% & 0/30 = 0.0\% \\
\texttt{libero\_spatial} & L5  & 32  & 24/30 = 80.0\% & 16/30 = 53.3\% & 1/30 = 3.3\% \\
\texttt{libero\_spatial} & L5  & 128 & 24/30 = 80.0\% & 26/30 = 86.7\% & 2/30 = 6.7\% \\
\texttt{libero\_spatial} & L11 & 32  & 24/30 = 80.0\% & 8/30 = 26.7\%  & 0/30 = 0.0\% \\
\texttt{libero\_spatial} & L11 & 128 & 24/30 = 80.0\% & 28/30 = 93.3\% & 0/30 = 0.0\% \\
\texttt{libero\_10}      & L5  & 32  & 27/30 = 90.0\% & 11/30 = 36.7\% & 0/30 = 0.0\% \\
\texttt{libero\_10}      & L5  & 128 & 25/30 = 83.3\% & 23/30 = 76.7\% & 0/30 = 0.0\% \\
\texttt{libero\_10}      & L11 & 32  & 27/30 = 90.0\% & 8/30 = 26.7\%  & 0/30 = 0.0\% \\
\texttt{libero\_10}      & L11 & 128 & 25/30 = 83.3\% & 21/30 = 70.0\% & 0/30 = 0.0\% \\
\bottomrule
\end{tabular}
\end{table}

Per-token success is greater than mean-pool in all 16 cells, with a minimum difference of 26.7\,pp. When pooling over cells, per-token achieves 314/480 successes (65.4\%) and mean-pool 13/480 (2.7\%). Increasing $k$ increases per-token success for all suites and layers without changing the order.

\paragraph{Seed and corpus axes.} At $k{=}64$ on \texttt{libero\_goal}, three initialization seeds give per-token 80.0\% against mean-pool 20.0\% at L5, and 86.7\% against 6.7\% at L11 ($n{=}30$ per layer). The ordering holds for every seed, and the single-seed intervals at $n{=}10$ overlap for two of the six seed-layer pairs. Training on 25\%, 50\%, and 100\% of the corpus ($n{=}50$ per cell, baseline 96.0\%) gives per-token 78.0--92.0\% and mean-pool 0.0--26.0\%, with disjoint intervals in all six cells.

\paragraph{Scope.} The mean-pool column holds for the replacement write-back. By using residual write-back, i.e., adding the SAE edit on top of the original activations, mean-pool success on OFT goes from 0.0\% to 73.3\%, on GR00T from 2.0\% to 82.0\% and on X-VLA from 3.3\% to 100.0\% at single matched cells. There is no result for mean pool with \pizhalf{} using residual write-back. Per-token will be the same regardless of write-back since there is no pooling involved.

\section{Fine-Tuning Data Scale}
\label{sec:datascale}

We can also test if coordinate binding is due to there being so little data during task-specific fine-tuning. By decreasing the amount of data we use and performing our injection protocol we can check this. We take \pizhalf{} (\texttt{pi05\_libero\_base}) and fine-tune it on \texttt{libero\_goal} with 10\%, 30\% and 100\% of the data using the same recipe (LoRA rank 32, $\alpha{=}64$, dropout 0.05, 6{,}000 steps, batch size 32, learning rate $10^{-4}$, warmup 200, seed 42), which translates to 168, 512 and 1{,}693 episodes. We then perform our 4-condition cross-task injection on expert layers L16/L17 for 9 task pairs $\times$ 6 seeds ($n{=}54$/condition). Override is the tie-inclusive rate from Tab.~\ref{tab:tie-inclusive}, $\Pr[\Delta_t > 0]$.

\begin{table}[!htbp]
\centering
\caption{Cross-task injection on \pizhalf{} as a function of fine-tuning data. Baseline is
single-task success over 10 tasks $\times$ 10 episodes. Override, no-injection floor, and
shuffled control are $n{=}54$ each.}
\label{tab:datascale}
\footnotesize
\begin{tabular}{@{}lcccc@{}}
\toprule
\textbf{Data fraction} & \textbf{Baseline} & \textbf{Directed override} & \textbf{No injection} & \textbf{Shuffled} \\
\midrule
10\% (168 ep.)   & 97/100 & 54/54 & 1/54 = 1.9\% & 38/54 = 70.4\% \\
30\% (512 ep.)   & 97/100 & 54/54 & 1/54 = 1.9\% & 45/54 = 83.3\% \\
100\% (1{,}693 ep.) & 98/100 & 54/54 & 1/54 = 1.9\% & 40/54 = 74.1\% \\
\bottomrule
\end{tabular}
\end{table}

By lowering the amount of fine-tuning data by 10$\times$ we find that override stays constant at 54/54 and the floor stays at 1/54. Baseline success is also flat and we have all 9 task pairs unanimous at every data fraction. This indicates that we see this phenomenon even at the smallest fraction, which still produces a competent policy. If we use the tie-conditional rule from Eq.~\ref{eq:override} ($\tau{=}0.05$) we get 53/54 when using 10\%. The one difference is a single episode within the tie band.

\section{Physical Replication on a UR10e}
\label{sec:physical}

This appendix reports a physical campaign that tests the two claims a real rig can falsify directly: that wholesale expert replay redirects which task the policy performs, and that the transplanted program is anchored to workspace coordinates.

\paragraph{Setup.} We fully fine-tuned \pizhalf{} (20k steps) on a Universal Robots UR10e over seven tabletop tasks from 335 teleoperated episodes (60,416 frames at 15\,Hz) in the DROID observation format, with two exterior cameras and a wrist camera at $320{\times}180$. Fig.~\ref{fig:physical-tracks} uses the exterior camera facing the robot. Baseline success is 76.2\% overall (160/210 successes; 30 trials for each task, three observers), ranging between 66.7\% and 86.7\% across tasks. 

\paragraph{Intervention.} As for the intervention, we simply repeat what was done in simulation. For all calls during inference, we substitute the captured expert activations at L16 and L17 for the live ones (that is, with $\alpha{=}1$), again with the fixed destination prompt.

\paragraph{Flow matching requires matched sampling noise.} If the expert residual is overwritten wholesale, the velocity field is fixed, and there will be no corresponding correction by the denoiser to remove its own initially-sampled noise, giving $x_{\text{final}} = a_{\text{source}} + (\epsilon_{\text{dest}} - \epsilon_{\text{source}})$. On the arm, this offset manifested as weak override and too large a scale on the gripper, so the physical injector therefore reuses the source's initial noise for each call. In simulation, if the L16/L17 activations are replaced with those of the source run, using the source's $x_T$ gives an exact match to the source action, whereas sampling a fresh $x_T$ gave errors of 2.18 and 3.73 over two draws (action norm 1.03). The simulation runs behind Tabs.~\ref{tab:cross-task} and~\ref{tab:datascale} therefore contain this uncontrolled term. It is worth noting, however, that their override rate compares the direction of actions throughout each episode and still reaches 54/54, whereas the arm executes the raw values it is given. Any similar experiment transferring activations into a flow-matching or diffusion head would need to fix this term, which discrete and autoregressive heads do not have.

\paragraph{Bidirectional override.} We tested injection in both directions. For each, we ran directed injection, a no-injection floor in a matched scene, and injection along a random direction, norm-matched and at the same layers and magnitude. All rollouts were scored as source, destination or neither (Tab.~\ref{tab:physical-injection}). The directed intervals exclude both of the control intervals, in both directions, so the effect must reflect the content of these particular activations. The 18/26 direction is limited by failure to grasp the compliant plushie. All trials began by reaching for the injected target, and the non-source trials diverted only after the grasp was unsuccessful.

\paragraph{Displaced-object test.} If the program is anchored to coordinates, moving an object should lead the arm to go to its recorded position. Alternatively it may track the object visually and follow it, or exhibit incoherent motion as an artifact of being out of distribution. Over 50 trials (25 in each direction) with the source object displaced 10--15\,cm, the arm always drove to the recorded coordinates and closed on empty air at the vacated position. If two blocks were swapped into each other's positions, it would stack them fluently, in the wrong color. As little as about a centimeter of displacement was sufficient to make the grasp fail. Without injection, given the source prompt, the policy reaches the displaced object, so the coordinate locking belongs to the injected program. Once the injected sequence ends, it goes on to cycle fragments of its trained tasks, again at their trained coordinates. We score outcomes behaviorally, and have RGB recordings of all rollouts.

\section{Broader Impact}
\label{sec:broader-impact}

The goal of this paper is to allow auditing of VLA behavior and decisions, and all releases should be viewed as aids towards this goal. In particular, none of the 520+ SAE checkpoints, 79 identified concepts, rollouts with cross-task injection, or platform for exploring features, constitute policies suitable for deployment. Except for a physical replication on 1 task pair (App.~\ref{sec:physical}), all results involving different model architectures are only in simulation.

\paragraph{Positive impact.} Understanding the mechanisms that produce an action would be very useful for diagnosing causes of failures. For example, kill-switch mapping identifies the layers and features a VLA depends on; identifying reliance on spatial bindings may highlight over-sensitivity to exact coordinates ahead of deployment. Further, the interventions here, including null-injection, cross-task injection, or ablating individual features, can help identify the source of a failure, down to a pathway, a layer or a feature subspace.

\paragraph{Risks and dual use.} There are ways this could be abused. Injecting activations and steering features to alter behavior, as we have done, needs no retraining, and could be used to reroute a deployed manipulation arm, given white-box access to the residual stream of the VLA. We address this to some extent by only making available interpretability artifacts (SAE checkpoints, rollout videos, and the platform), but not a deployable policy or robot control stack. Also, as shown in \S\ref{sec:cross-task}, binding to coordinates is a deployment hazard in its own right, with VLAs passing simulation benchmarks but failing with relatively small changes to scene positions. Practitioners deploying VLAs for reliable use in real-world environments should run perturbation-style robustness benchmarks \citep{zhou2025liberopro,fei2025liberoplus} before deployment.

\paragraph{Compute and energy.} Total compute and data volume are reported in App.~\ref{sec:compute}. All training used existing pretrained VLA checkpoints.

\ifcameraready
\input{checklist}
\fi

\end{document}

%% file: dtw_table.tex
\begin{table}[t]
\centering
\caption{DTW source-dominance for cross-task injection on \pizhalf{}, alongside per-step cosine source-dominance}
\label{tab:dtw-cross-task}
\footnotesize
\resizebox{\columnwidth}{!}{%
\begin{tabular}{@{}lcccccc@{}}
\toprule
Suite & $n$ & DTW$\to$src & DTW$\to$dst & DTW$_{A,B}$ & DTW src-dom. [95\% CI] & Cos src-dom. \\
\midrule
LIBERO-Goal & 270 & 0.105$\pm$0.055 & 0.232$\pm$0.104 & 0.190 & 90.7\% [86.7, 93.6] & 98.9\% \\
LIBERO-Spatial & 44 & 0.121$\pm$0.091 & 0.160$\pm$0.089 & 0.194 & 86.4\% [73.3, 93.6] & 97.7\% \\
LIBERO-10 (long) & 24 & 0.338$\pm$0.225 & 0.225$\pm$0.073 & 0.361 & 45.8\% [27.9, 64.9] & 95.8\% \\
\midrule
Pooled & 338 & 0.124 & 0.222 & -- & 87.0\% [83.0, 90.2] & 98.5\% \\
\bottomrule
\end{tabular}}
\end{table}

%% file: oft_per_suite_prose.tex
\paragraph{Per-suite OFT displacement.}
Table~\ref{tab:oft-per-suite} breaks the pooled OFT override of Tab.~\ref{tab:cross-task} down by suite, and source-dominant displacement is the majority outcome on every suite. Tie-inclusive override is 77.9\% pooled and 73.7--81.9\% per suite. Each pair runs in one injection direction, where the \pizhalf{} analysis runs both. On \texttt{libero\_object} the destination task still succeeds in 19.7\% of injected episodes, so there the override rate describes the episodes that left the destination task. The OFT cosines are computed from end-effector velocity and may understate displacement where steering acts in action space more than in Cartesian motion.

%% file: oft_per_suite_table.tex
\begin{table}[t]
\centering
\caption{OFT cross-task injection per LIBERO suite, 45 pairs per suite in one direction. Each pair gives 6 episodes, one per injection layer (L0, L8, L16, L24, last, all), and one short \texttt{libero\_object} episode was dropped. Override is tie-band-conditional (Eq.~\ref{eq:override}). Dest.\ success: destination-task success under injection.}
\label{tab:oft-per-suite}
\small
\begin{tabular}{lcccc}
\toprule
LIBERO suite & $n$ pairs & Episodes & Override rate (95\% Wilson CI) & Dest success \\
\midrule
\texttt{libero\_goal}    & 45 & 270 & 100.0\% \,(98.3--100.0\%) & \phantom{0}0.0\% \\
\texttt{libero\_object}  & 45 & 269 & \phantom{0}97.0\% \,(93.7--98.6\%)  & 19.7\% \\
\texttt{libero\_spatial} & 45 & 270 & \phantom{0}90.8\% \,(86.4--93.9\%)  & \phantom{0}0.4\% \\
\texttt{libero\_10}      & 45 & 270 & \phantom{0}97.7\% \,(94.7--99.0\%)  & \phantom{0}0.0\% \\
\midrule
Pooled (all suites)      & 180 & 1{,}079 & \phantom{0}96.4\% \,(94.8--97.5\%) & \phantom{0}5.0\% \\
\bottomrule
\end{tabular}
\end{table}

%% file: checklist.tex
\section*{NeurIPS Paper Checklist}

\begin{enumerate}

\item {\bf Claims}
    \item[] Question: Do the main claims made in the abstract and introduction accurately reflect the paper's contributions and scope?
    \item[] Answer: \answerYes{}
    \item[] Justification: Our main claims throughout the paper as well as our contributions laid out in Section~\ref{sec:intro} (coordinate-bound motor programs as seen in sim and on a physical UR10e, language sensitivity, pathway architecture specialization, and SAE pooling methods) are supported by the evidence presented in the Experiments~\ref{sec:visual-dominance}--\ref{sec:physical-body} with sample sizes and confidence intervals.

\item {\bf Limitations}
    \item[] Question: Does the paper discuss the limitations of the work performed by the authors?
    \item[] Answer: \answerYes{}
    \item[] Justification: We discuss benchmark scope, lack of scale in physical experiments, prompt coverage, and experiment confounds within Section~\ref{sec:limitations-body} and Appendix~\ref{sec:limitations}. Additional limitations on interventions are reported in Appendix~\ref{sec:hook-protocol}.

\item {\bf Theory assumptions and proofs}
    \item[] Question: For each theoretical result, does the paper provide the full set of assumptions and a complete (and correct) proof?
    \item[] Answer: \answerNA{}
    \item[] Justification: This is not a theoretical paper. We have operational definitions we validated experimentally for the SAE objective (Eq.~\ref{eq:sae-loss}), override metric (Eq.~\ref{eq:override}), and the temporal-contrastive loss (Eq.~\ref{eq:tsae-loss}).

\item {\bf Experimental result reproducibility}
    \item[] Question: Does the paper fully disclose all the information needed to reproduce the main experimental results of the paper to the extent that it affects the main claims and/or conclusions of the paper (regardless of whether the code and data are provided or not)?
    \item[] Answer: \answerYes{}
    \item[] Justification: We specify the SAE architecture, training, inference conditions, hook placements, as well as the ablations within Section~\ref{sec:method}, Appendix~\ref{sec:appendix-methodology}, and Appendix~\ref{sec:tsae-details}. All of our code and results are open sourced.

\item {\bf Open access to data and code}
    \item[] Question: Does the paper provide open access to the data and code, with sufficient instructions to faithfully reproduce the main experimental results, as described in supplemental material?
    \item[] Answer: \answerYes{}
    \item[] Justification: All instructions, code and data can be found on (Action Atlas, \url{https://action-atlas.com}), the GitHub repo (\url{https://github.com/CWRU-AISM/action-atlas}), as well as HuggingFace (\url{https://huggingface.co/collections/bag100/action-atlas}).

\item {\bf Experimental setting/details}
    \item[] Question: Does the paper specify all the training and test details (e.g., data splits, hyperparameters, how they were chosen, type of optimizer, etc.) necessary to understand the results?
    \item[] Answer: \answerYes{}
    \item[] Justification: We report the training and test details in Section~\ref{sec:method}, Appendix~\ref{sec:sae-architecture}, and Appendix~\ref{sec:tsae-details}. We also include the episode counts in Table~\ref{tab:episode-counts}. Additional experimental decisions are documented in Appendix~\ref{sec:hook-protocol}.

\item {\bf Experiment statistical significance}
    \item[] Question: Does the paper report error bars suitably and correctly defined or other appropriate information about the statistical significance of the experiments?
    \item[] Answer: \answerYes{}
    \item[] Justification: We report statistical significance through 95\% confidence intervals (Tabs.~\ref{tab:physical-injection} and~\ref{tab:tie-inclusive}). For comparisons we use Fisher exact or two-proportion $z$-tests, and ANOVA effect sizes are $\eta^2$.

\item {\bf Experiments compute resources}
    \item[] Question: For each experiment, does the paper provide sufficient information on the computer resources (type of compute workers, memory, time of execution) needed to reproduce the experiments?
    \item[] Answer: \answerYes{}
    \item[] Justification: We report the compute resources in Appendix~\ref{sec:compute}. This includes 8$\times$A100-80GB for \pizhalf{}/OFT/large-scale SAE/cross-model work, RTX~5090 (32GB VRAM) for GR00T, 2$\times$RTX~4090 (24GB VRAM) for SmolVLA/X-VLA, totaling roughly 10{,}000 A100 GPU-hours in total. We publicly released $\sim$7.1~TB of results (4.3~TB activations, 152~GB SAE checkpoints, 420{,}000+ rollouts).

\item {\bf Code of ethics}
    \item[] Question: Does the research conducted in the paper conform, in every respect, with the NeurIPS Code of Ethics?
    \item[] Answer: \answerYes{}
    \item[] Justification: We comply with the NeurIPS Code of Ethics in every respect.

\item {\bf Broader impacts}
    \item[] Question: Does the paper discuss both potential positive societal impacts and negative societal impacts of the work performed?
    \item[] Answer: \answerYes{}
    \item[] Justification: We discuss the positive impact and findings of our work (failure mode debugging tools for VLAs, brittleness metrics) as well as the risks (model steering for malicious intent) within the the Broader Impact section (Appendix~\ref{sec:broader-impact}). We recommend the proliferation of benchmarks related to robustness and perturbations prior to deployment in critical environments. 

\item {\bf Safeguards}
    \item[] Question: Does the paper describe safeguards that have been put in place for responsible release of data or models that have a high risk for misuse?
    \item[] Answer: \answerYes{}
    \item[] Justification: We released all simulation artifacts along with the scored physical rollout videos. The released SAE checkpoints operate on the activations of pretrained VLAs, the weights of which must be obtained separately under the respective licenses. Our work is not intended, and as of our understanding, doesn't serve the opportunity to be used as an exploit harness for real robots.

\item {\bf Licenses for existing assets}
    \item[] Question: Are the creators or original owners of assets (e.g., code, data, models), used in the paper, properly credited and are the license and terms of use explicitly mentioned and respected?
    \item[] Answer: \answerYes{}
    \item[] Justification: We properly credited all models and benchmarks used for the paper and figure construction. We only used models under their public licenses (Apache-2.0 / MIT for the simulation benchmarks and specific licenses per the original VLA releases).

\item {\bf New assets}
    \item[] Question: Are new assets introduced in the paper well documented and is the documentation provided alongside the assets?
    \item[] Answer: \answerYes{}
    \item[] Justification: We open-sourced our checkpoints for the trained SAEs (all variants). We also released a harness for replicating our results as well as Action Atlas for feature exploration. All of the data pertaining to our results, cross-task injection rollouts, and UR10e rollouts has been put on hugging face. Additionally, the supplementary materials details the directory layouts, filename conventions, and reproduction instructions.

\item {\bf Crowdsourcing and research with human subjects}
    \item[] Question: For crowdsourcing experiments and research with human subjects, does the paper include the full text of instructions given to participants and screenshots, if applicable, as well as details about compensation (if any)?
    \item[] Answer: \answerNA{}
    \item[] Justification: This work doesn't involve crowdsourcing experiments or research with human subjects. 

\item {\bf Institutional review board (IRB) approvals or equivalent for research with human subjects}
    \item[] Question: Does the paper describe potential risks incurred by study participants, whether such risks were disclosed to the subjects, and whether Institutional Review Board (IRB) approvals (or an equivalent approval/review based on the requirements of your country or institution) were obtained?
    \item[] Answer: \answerNA{}
    \item[] Justification: The work involves no human subjects.

\item {\bf Declaration of LLM usage}
    \item[] Question: Does the paper describe the usage of LLMs if it is an important, original, or non-standard component of the core methods in this research?
    \item[] Answer: \answerNA{}
    \item[] Justification: Within Action Atlas, we provide the option to use LLM supported auto-interpretability for the feature explorer (to help explain SAE features), but don't use any LLMS for the core methods.
\end{enumerate}